\documentclass{article}

\usepackage[preprint]{corl_2026} 

\usepackage{amsmath}
\usepackage{amsfonts}
\usepackage{graphicx}
\usepackage{adjustbox}

\usepackage{multirow}
\usepackage{makecell}
\usepackage{xspace}
\usepackage{stfloats}
\usepackage{float}
\usepackage[dvipsnames]{xcolor}

\newcommand{\tavla}{TA-VLA\xspace}
\newcommand{\forcevla}{ForceVLA\xspace}
\newcommand{\method}{\textbf{FACT}\xspace}
\newcommand{\methodpizero}{\textbf{FACT}$_{\pi_0}$\xspace}
\newcommand{\methodpizeroft}{\textbf{FACT}$_{\pi_0\text{.5}}$\xspace}

\newcommand{\placeholder}[1]{\textcolor{lightgray}{--}}

\usepackage{colortbl}  %
\definecolor{lavender}{rgb}{0.9, 0.9, 0.98}
\definecolor{red}{RGB}{255, 0, 0}
\definecolor{orange}{RGB}{252, 130, 62}
\definecolor{blue}{RGB}{0, 0,255}
\definecolor{darkgreen}{RGB}{0, 180,0}
\definecolor{lightgray}{gray}{0.9}
\definecolor{lightblue}{rgb}{0.68,0.85,0.9}
\definecolor{lightorange}{rgb}{1.0, 0.8, 0.6}
\definecolor{myblue}{RGB}{30, 100, 200}

\newcommand{\note}[1]{\color{orange}}
\usepackage{flafter}
\usepackage{afterpage}
\usepackage{booktabs}
\usepackage{tabularx}
\usepackage{subcaption}

\definecolor{factblue}{RGB}{30,100,200}
\definecolor{rowyellow}{HTML}{FFF5E6}
\definecolor{facttau}{RGB}{90,40,170}
\definecolor{factforce}{RGB}{200,95,20}

\title{\textbf{Demystifying When and Why VLAs Fail in Contact-Rich Tasks and How to Fix Them}}

\author{
  Carlota Parés-Morlans, Nils Kuhn, Y. Isabel Liu, Alberta Longhini, Jeannette Bohg\\
  Stanford University\\\\
  \href{https://stanford-iprl-lab.github.io/fact/}{https://stanford-iprl-lab.github.io/fact/}\\
}

\begin{document}
\maketitle
\vspace{-1em}

\begin{abstract}
We address the problem of understanding when and why Vision-Language-Action models struggle with contact-rich manipulation tasks that require precise physical interaction. Prior work has primarily focused on addressing contact failures through force-augmented architectures and training-time regularizers, yet the factors contributing to these failures remain underexplored. We identify two distinct failure modes underlying this gap. Precision failures occur when policies struggle to produce the small, accurate corrective actions required during contact and force failures occur when policies fail to respond to contact forces. To address these failures, we propose \method. During post-training, \method increases low noise sampling to improve fine action prediction and incorporates structured force conditioning to guide action generation using contact state and force history. \method achieves an average success rate of 66\% across five contact-rich tasks against 41\% for the best prior baseline, in an evaluation spanning almost 2{,}500 real-world rollouts.%
\end{abstract}


\section{Introduction}
\label{sec:intro}

Vision-Language-Action (VLA) models have significantly advanced robotic manipulation, enabling policies that generalize across diverse tasks such as object retrieval and shirt folding~\citep{black2024pi0,black2025pi05,zitkovich2023rt,bjorck2025gr00t,kim2024openvla,shukor2025smolvla,o2024open}. Despite these advances, contact-rich tasks such as connector insertion and precision assembly remain an open challenge. Unlike free-space manipulation, these tasks require continuous force-regulated interaction where contact forces vary with part geometry, material properties, and surface compliance. At the same time, visual feedback often degrades due to self-occlusion at the moment when precise corrective actions are most critical. Standard training paradigms and architectural choices are ill-suited to these requirements, and current models continue to struggle in this regime.

\begin{figure*}[t]
  \centering
  \setlength{\tabcolsep}{1pt}
  \begin{tabular}{ccccc}
    \includegraphics[width=0.19\linewidth]{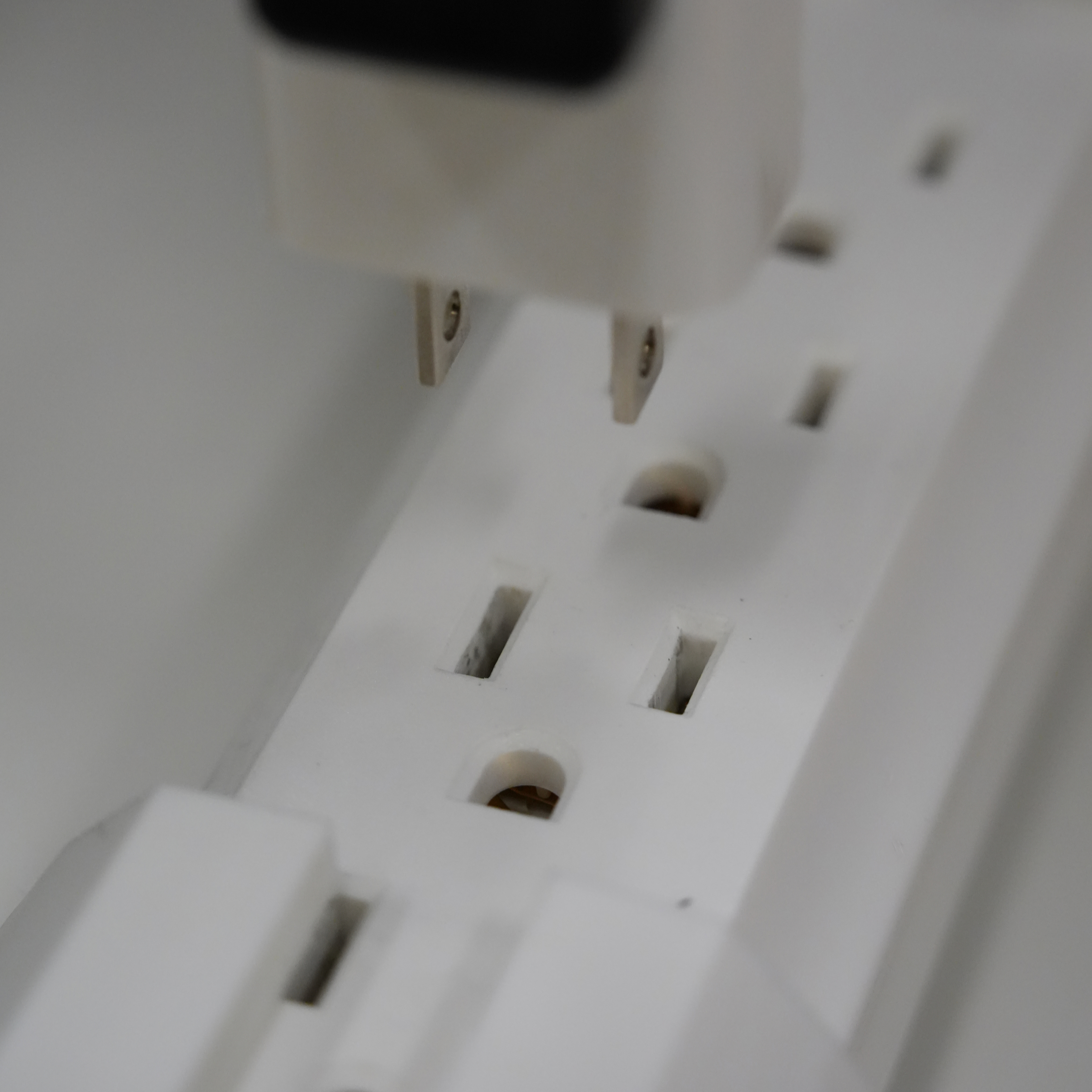} &
    \includegraphics[width=0.19\linewidth]{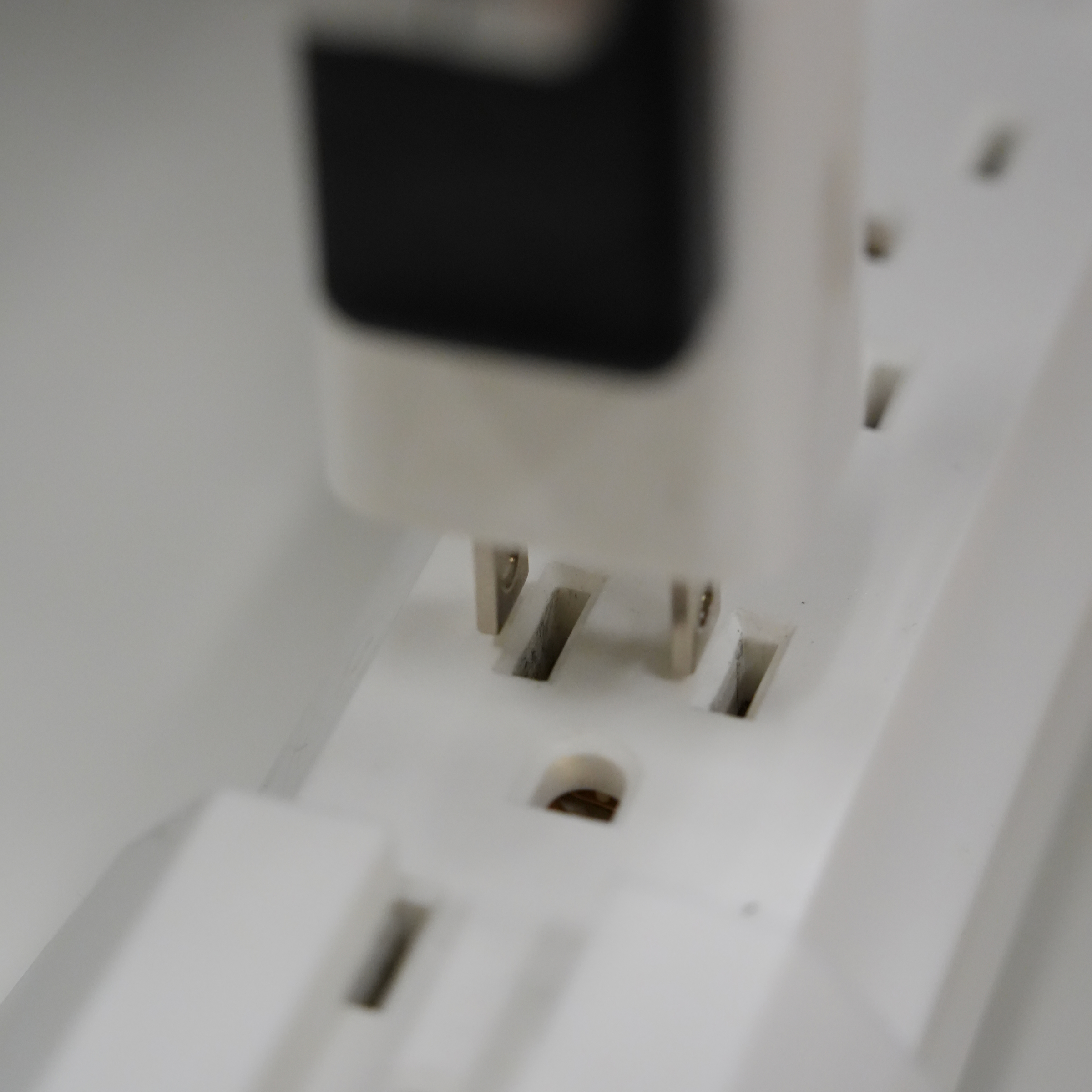} &
    \includegraphics[width=0.19\linewidth]{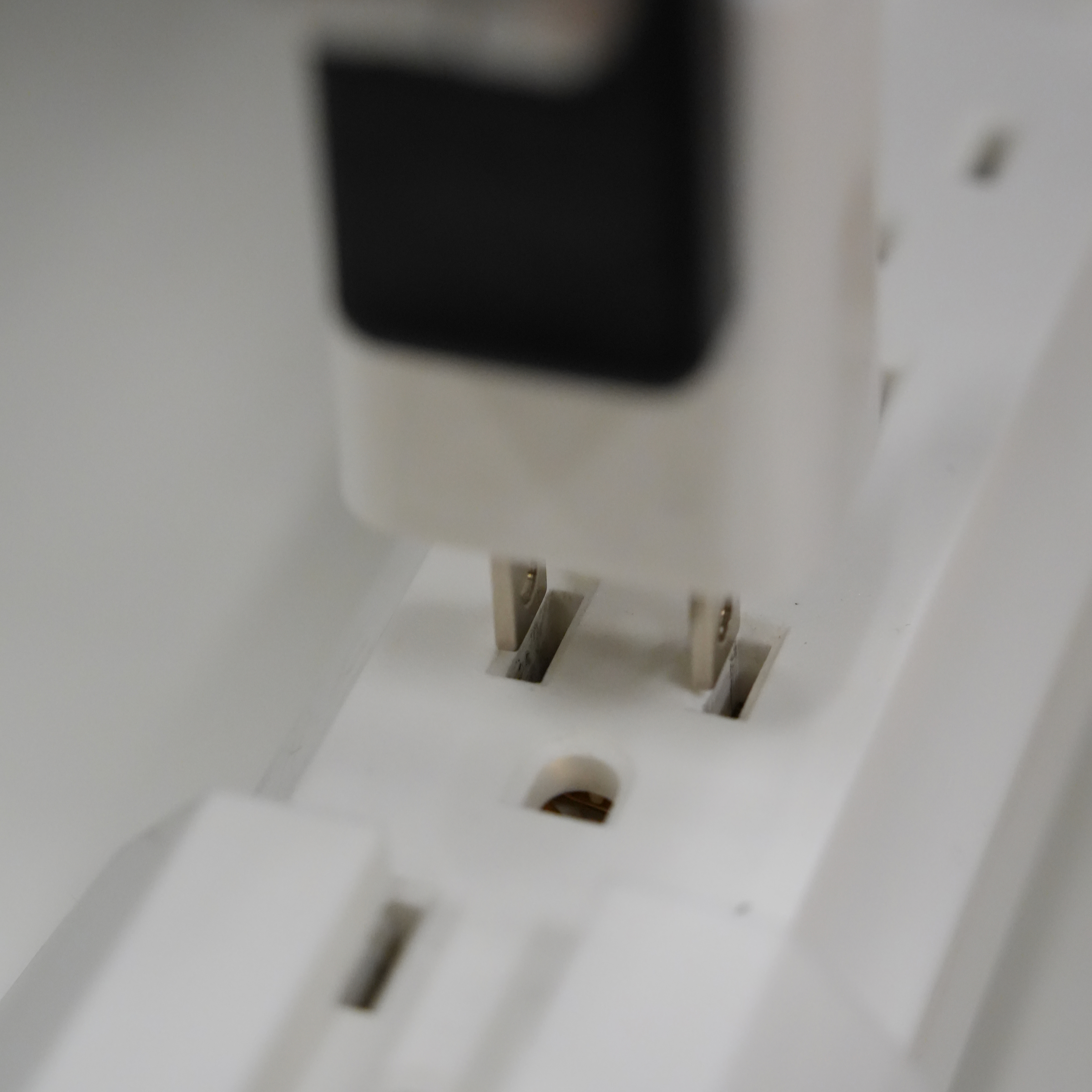} &
    \includegraphics[width=0.19\linewidth]{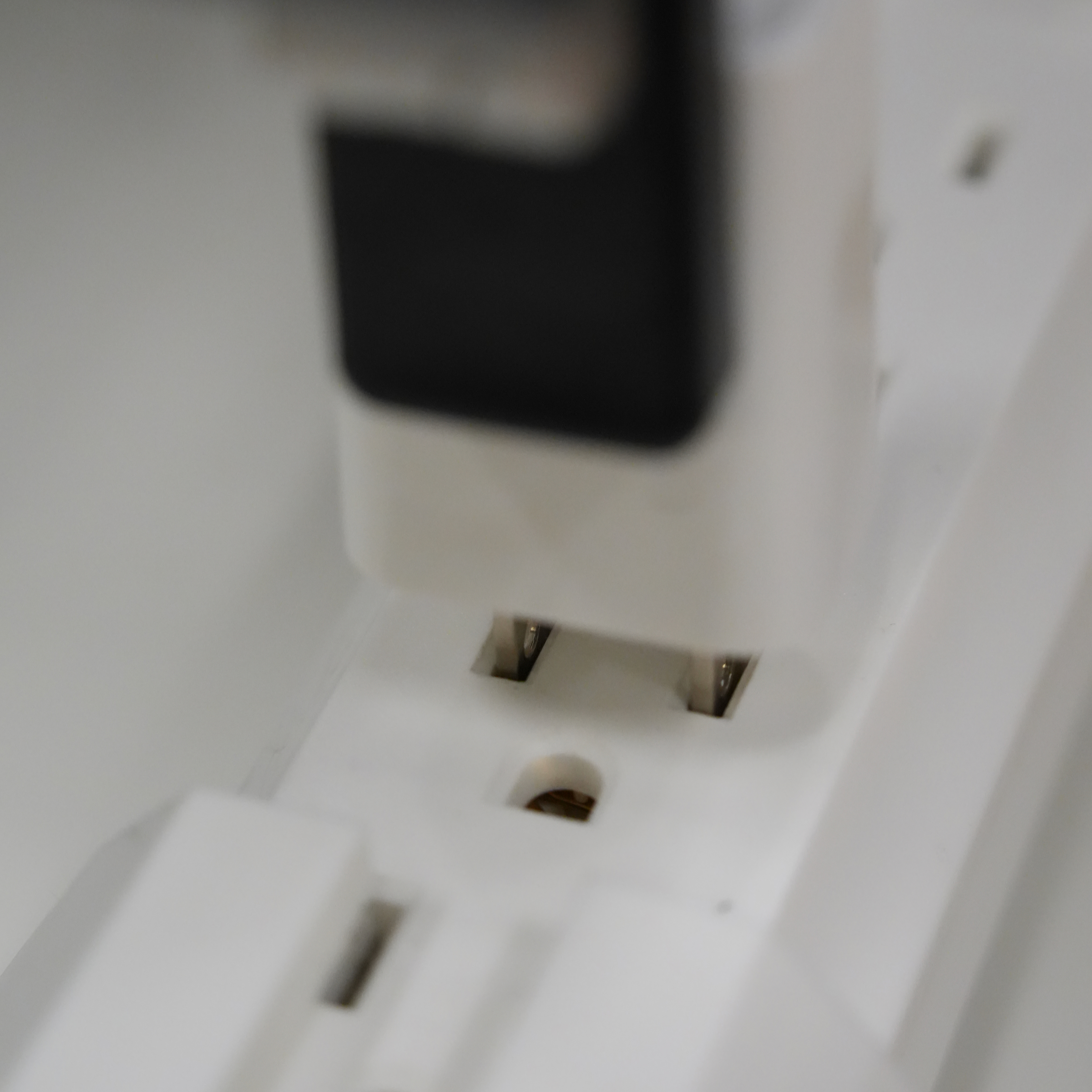} &
    \includegraphics[width=0.19\linewidth]{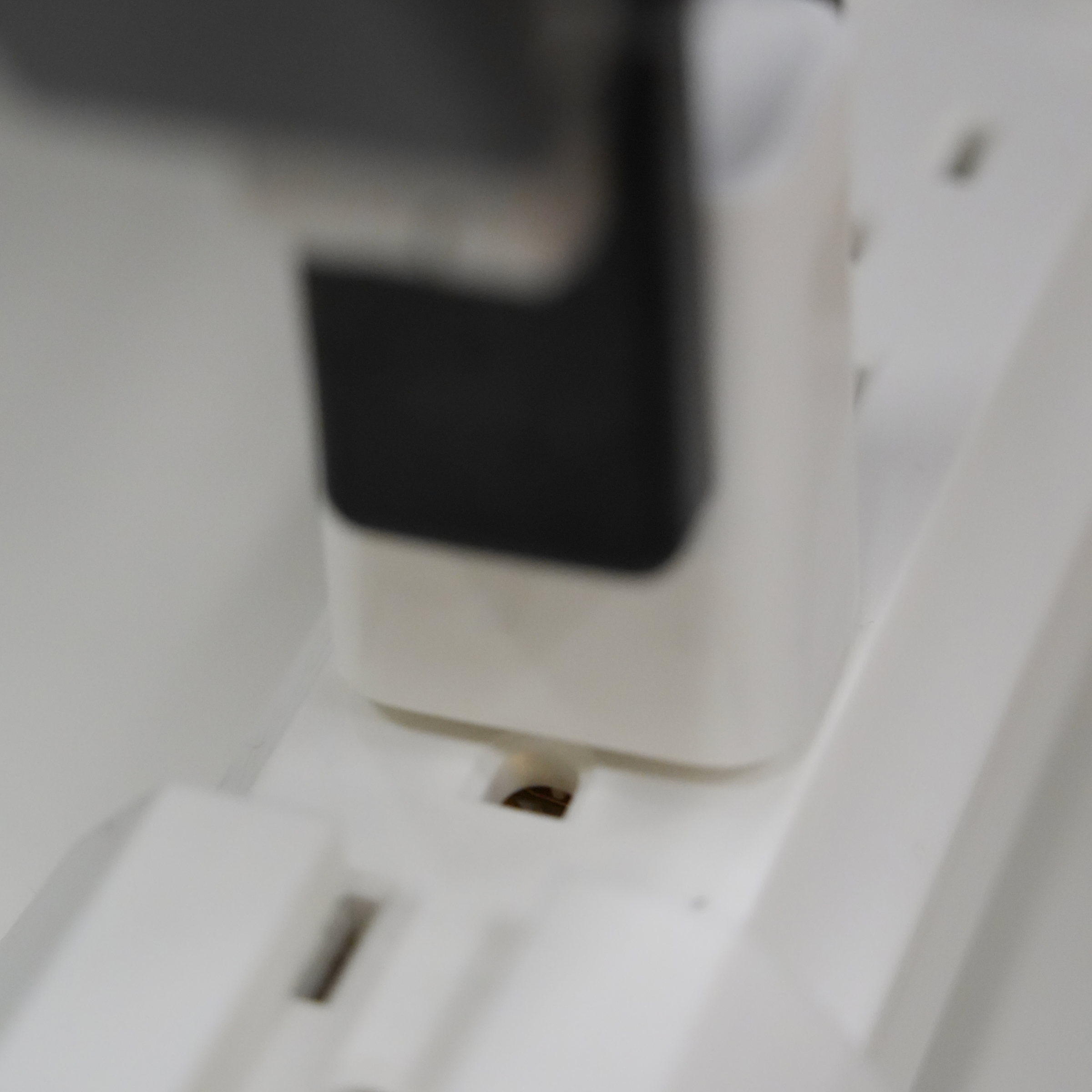} \\[-1pt]
    \includegraphics[width=0.19\linewidth]{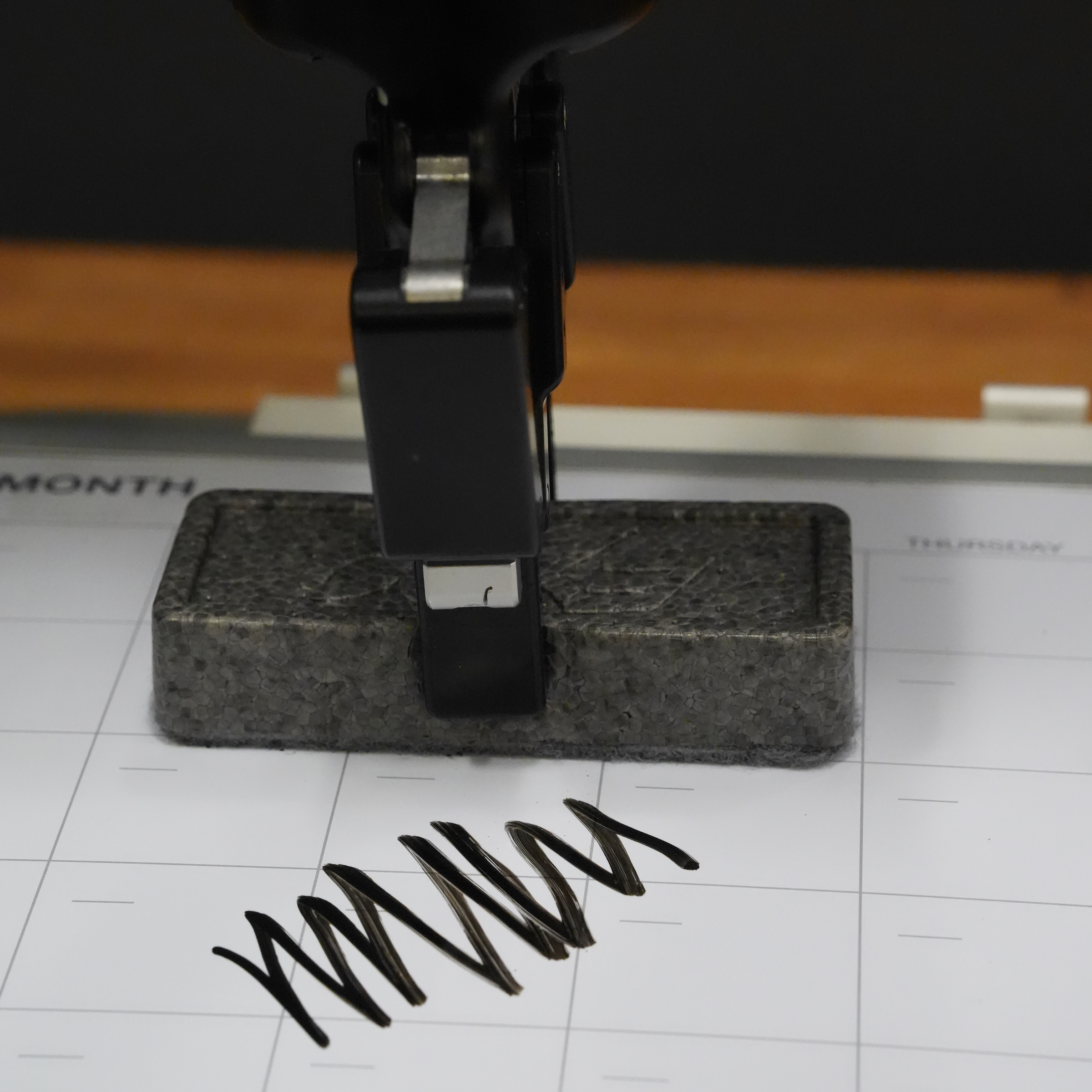} &
    \includegraphics[width=0.19\linewidth]{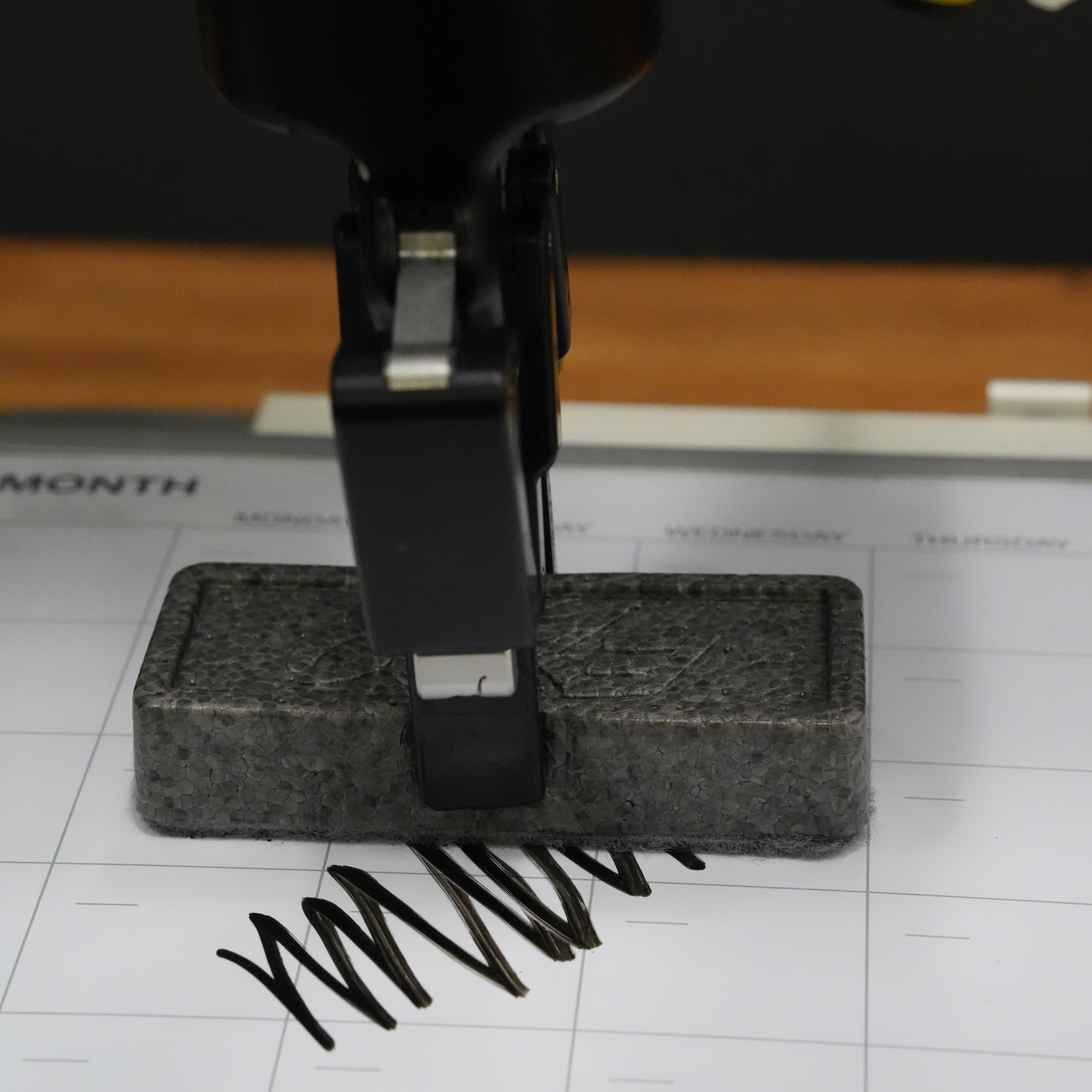} &
    \includegraphics[width=0.19\linewidth]{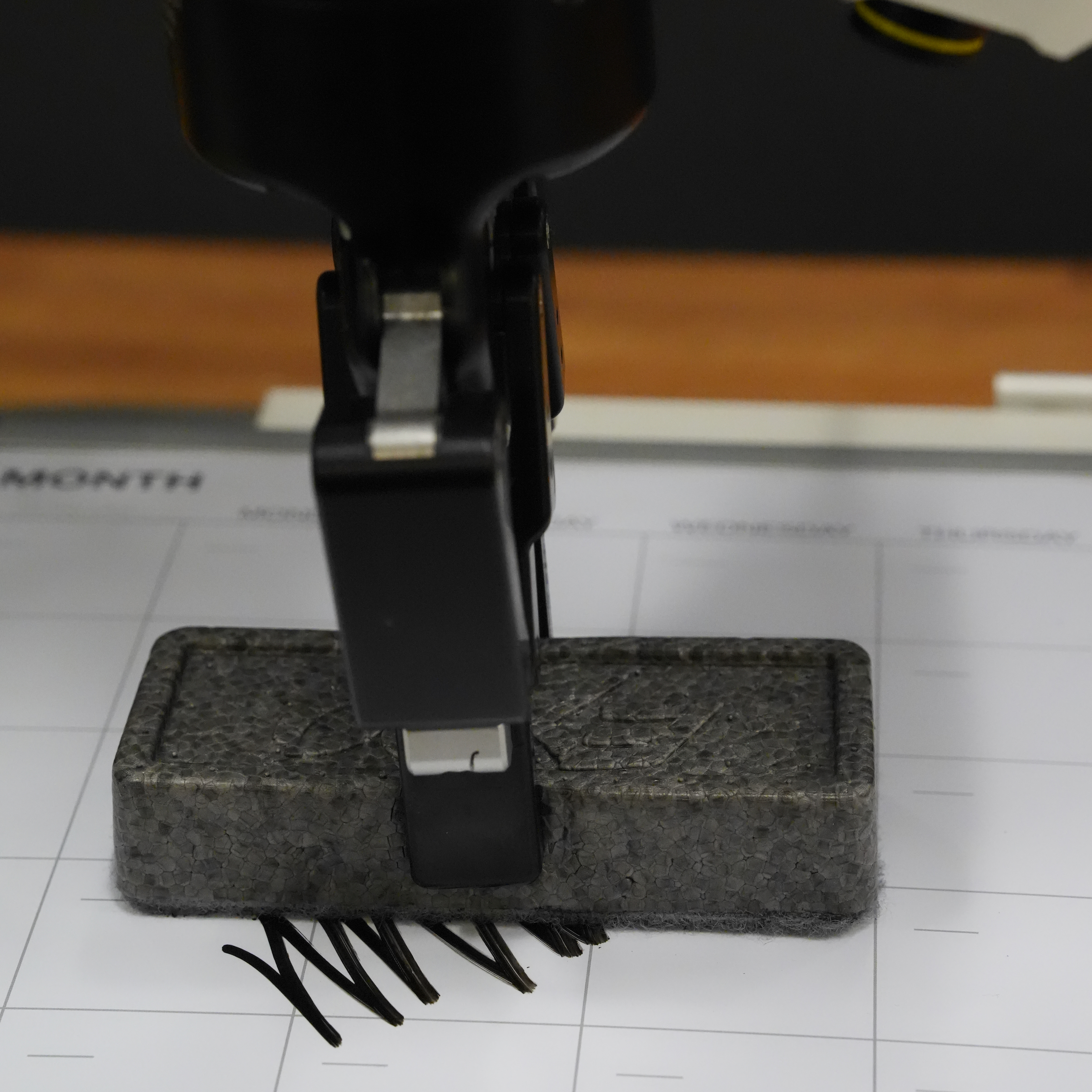} &
    \includegraphics[width=0.19\linewidth]{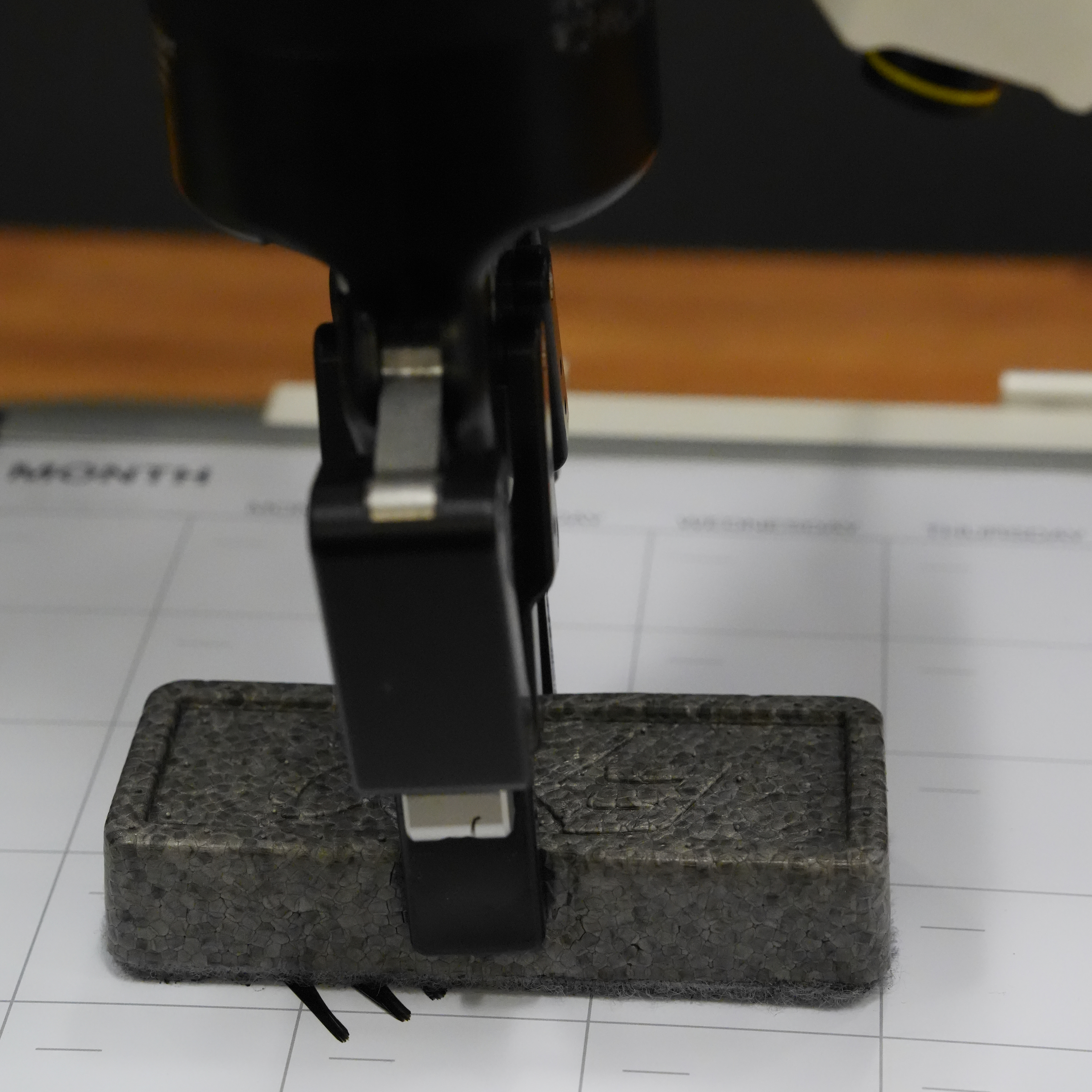} &
    \includegraphics[width=0.19\linewidth]{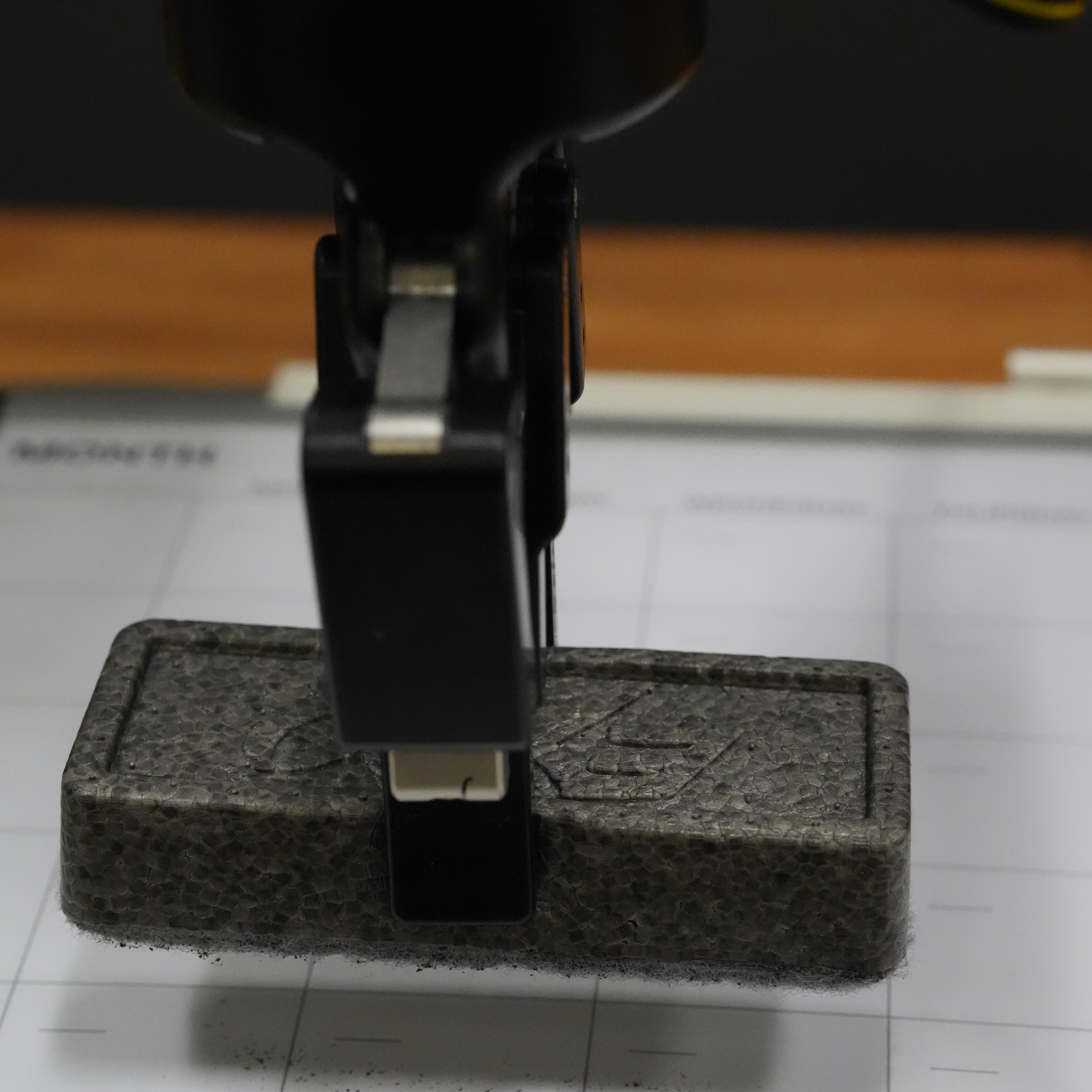} \\
  \end{tabular}
  \caption{\textbf{Contact-rich manipulation tasks from our experiments.} These tasks involve sustained or repeated physical interaction, where success depends on simultaneously regulating both motion and contact forces measured by a wrist-mounted force/torque sensor. Plug insertion (top) demands sub-millimeter alignment of the connector under partial occlusion while managing contact until the connector is fully seated in the socket. Board erasing (bottom) requires maintaining consistent contact with the surface throughout the stroke without damaging the surface.
  }
  \vspace{-10pt}
  \label{fig:teaser}
\end{figure*}

A growing line of work attempts to close this contact-rich manipulation gap by augmenting VLA architectures with force feedback, either through dedicated fusion modules~\citep{yu2025forcevla,tavla2025,li2026favla} or by using force signals as auxiliary supervision during training~\citep{zhang2026craft,zhao2026fdvla}. These approaches share two implicit assumptions. First, they frame the contact-rich manipulation gap primarily as a force-sensing problem, leaving unexamined whether failures could stem from the training procedure itself rather than from missing sensory modalities. Second, when force is incorporated, it is usually appended alongside vision and proprioception, which may limit how effectively the policy exploits the properties of force signals.

We diagnose why VLA policies often fail in contact-rich tasks and identify two distinct failure modes, \emph{precision} and \emph{force failures}. Precision failures arise when the policy struggles to produce the small, accurate corrective actions required for contact-rich manipulation. Force failures occur because force signals contain informative dynamics that are not captured when force is treated as just another input stream. To address both failure modes, we propose \method, Force-Aware Contact-Rich Manipulation via Timestep Modulation. \method combines a noise schedule that emphasizes low-noise timesteps during post-training with time-aware force conditioning that integrates contact state and force history.

We evaluate \method across five contact-rich tasks, two of which are illustrated in Figure~\ref{fig:teaser}, spanning both precision-driven and force-driven failure modes. \method consistently outperforms force-augmented VLA baselines by more than 25 percentage points. Ablations show that each component contributes to these gains. The targeted noise schedule improves performance by up to 20 percentage points on precision-critical tasks, while time-aware force injection adds up to 22.5 percentage points on force-critical tasks.

In summary, this paper makes the following contributions:
\begin{itemize}
  \item A principled decomposition of VLA failures in contact-rich tasks into two modes, \emph{precision failures} and \emph{force failures.}

  \item A targeted noise schedule that addresses precision failures by increasing sampling of low-noise timesteps, improving precision-task success on the tested flow-matching VLA backbones, with no additional parameters, data, or architectural changes.

  \item A time-aware force injection mechanism that addresses force failures by integrating force signals according to their distinctive properties.
\end{itemize}

\section{Related Work}
\label{sec:related}

\subsection{Contact-Rich Manipulation}

Contact-rich manipulation comprises tasks whose success depends not only on reaching a goal pose, but also on reasoning through physical interaction with the environment, including friction, compliance, jamming, and force regulation. This diversity motivates distinct sensing and control strategies across the literature. Classical approaches rely on haptic feedback alone, where passive mechanical compliance shapes the wrench response to misalignment~\citep{whitney1982}, active force control regulates end-effector wrenches during contact~\citep{khatib1987,lefebvre2005}, and search strategies resolve position uncertainty through structured exploration~\citep{chhatpar2001}. Learning-based methods extend haptics-only control by learning contact-aware behaviors directly from interaction. For example, FORGE uses reinforcement learning to solve peg insertion, gear meshing, and nut threading from wrist force alone~\citep{noseworthy2025forge}. Adding vision unlocks spatial reasoning and generalization. Vision-force methods fuse wrist-mounted or fingertip force sensing with visual observations for insertion under occlusion, surface wiping, and force-sensitive manipulation~\citep{lee2020makingsense,he2025foar,li2025hybrid,chen2025dexforce}, and vision-tactile methods combine visual observations with distributed fingertip sensors for dexterous contact-rich tasks~\citep{bi2025vla}. Combining vision with haptic sensing is non-trivial, as haptic signals are sparse and localized while vision is dense and structured, and naive fusion often overfits to the visual stream~\citep{liu2025factr}. Our work builds on this fusion literature but focuses on exploiting the properties of force signals to propose a simple and effective mechanism for integrating them into vision-based models.

\subsection{Force-augmented Vision-Language-Action Models}

Vision-language-action models~\citep{black2024pi0,black2025pi05,zitkovich2023rt,bjorck2025gr00t,kim2024openvla,shukor2025smolvla,o2024open} have rapidly become the dominant paradigm for generalist manipulation, yet they still underperform on contact-rich tasks. A growing line of work has sought to close this performance gap by augmenting VLAs with force feedback, addressing three distinct challenges. The first is ensuring force reliably shapes action selection during contact, with approaches ranging from force-aware routing after the VL encoder~\citep{yu2025forcevla} to torque-history tokens~\citep{tavla2025} and per-layer cross-attention~\citep{li2026favla}. The second is action-space resolution, where fine-grained contact corrections require faster updates than a VLM backbone can provide, with approaches including decoupling a slow VLM backbone from a fast action module~\citep{li2026favla}. The third is modality imbalance, preventing force from being dominated by the visual stream, with approaches ranging from information bottlenecks~\citep{zhang2026craft} to training-time distillation~\citep{zhao2026fdvla}. In contrast to prior work that addresses contact failures through architectural force integration, we show that the training noise schedule and structural properties of force signals play an equally important role, and evaluate our approach, \method, against representative force-augmented VLAs, ForceVLA and TA-VLA~\citep{yu2025forcevla,tavla2025}.

\section{Preliminaries}
\label{sec:preliminaries}

We introduce the flow-matching policy formulation and notation used throughout the paper, with emphasis on the role of the noise level $\tau$ and its sampling distribution.

Flow matching~\citep{lipman2022flow} trains a network $v_\theta$ to predict a time-dependent velocity field that maps samples from a noise distribution to a target distribution.
In visuomotor policy learning, the target is the conditional distribution of action chunks given the current observation $\mathbf{o}$.
Given a clean action chunk $\mathbf{a}^0$ and noise $\boldsymbol{\epsilon}\sim\mathcal{N}(\mathbf{0},I)$, the rectified linear interpolant defines noisy actions $\mathbf{a}^\tau = (1-\tau)\mathbf{a}^0 + \tau\boldsymbol{\epsilon}$ for $\tau\in[0,1]$, with constant target velocity $\boldsymbol{\epsilon}-\mathbf{a}^0$.
The network $v_\theta$ is trained to match this target, conditioned on the observation $\mathbf{o}$ and noise level $\tau$, via the flow-matching objective
\begin{equation}
    \mathcal{L}_{\mathrm{FM}}(\theta)
    =
    \mathbb{E}_{\mathbf{a}^0,\boldsymbol{\epsilon},\tau}
    \left[
    \left\|
    v_{\theta}(\mathbf{o}, \mathbf{a}^{\tau}, \tau)
    -
    (\boldsymbol{\epsilon}-\mathbf{a}^0)
    \right\|_2^2
    \right].
\end{equation}

At inference time, integrating the learned field from $\tau{=}1$ to $\tau{=}0$ maps an initial Gaussian noise action chunk $\mathbf{a}^1$ to a clean action chunk $\mathbf{a}^0$. The training distribution of $\tau$ is set by a noise-level scheduler. Many standard choices, including the beta schedulers used in recent flow-based VLAs such as $\pi_{0.5}$~\citep{black2025pi05}, concentrate probability mass on large $\tau$ values giving the near-clean regime fewer training samples. We focus our evaluation primarily on $\pi_{0.5}$, with additional results on $\pi_0$~\citep{black2024pi0} and GR00T N1~\citep{bjorck2025gr00t} (Section~\ref{sec:results}).

\section{Diagnosing Failures of VLA Policies in Contact-Rich Tasks}

\label{sec:demistifying_failures}

\begin{figure*}[ht!]
\centering

\newcommand{\axshift}{70pt}

\begin{subfigure}[b]{0.32\textwidth}
\centering
\resizebox{\linewidth}{!}{\includegraphics{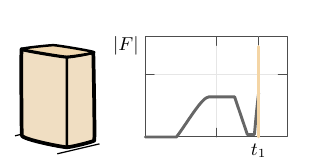}}
\caption{Success}
\label{fig:plug-insertion-a}
\end{subfigure}
\hfill
\begin{subfigure}[b]{0.32\textwidth}
\centering
\resizebox{\linewidth}{!}{\includegraphics{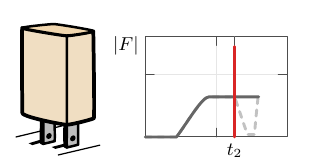}}
\caption{Precision failure}
\label{fig:plug-insertion-b}
\end{subfigure}
\hfill
\begin{subfigure}[b]{0.32\textwidth}
\centering
\resizebox{\linewidth}{!}{\includegraphics{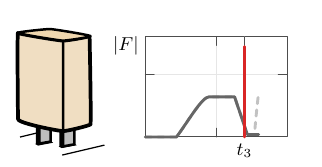}}
\caption{Force failure}
\label{fig:plug-insertion-c}
\end{subfigure}

\caption{\textbf{VLA failure modes on plug insertion.}
  Force magnitude $|F|$ is shown over time. Dashed grey traces repeat the success profile for reference.
  \textbf{(a)}~\emph{Success}. The plug approaches the socket ($|F|{=}0$), makes contact and explores alignment ($|F|{>}0$), enters the socket ($|F|{\approx}0$), and fully seats, producing the sharp force rise at $t_1$.
  \textbf{(b)}~\emph{Precision failure}. The plug is misaligned at the socket entry and force builds and saturates at $t_2$ as the plug presses against the rim.
  \textbf{(c)}~\emph{Force failure}. The plug enters the socket correctly but the policy stops short of full insertion and no seating force rise is detected at $t_3$.
  These examples illustrate distinct failure behaviors that motivate different corrective strategies.}
  \vspace{-10pt}
\label{fig:plug-insertion}
\end{figure*}

Compared to general manipulation tasks, contact-rich tasks require
both fine-grained correction and reasoning about contact forces.
Figure~\ref{fig:plug-insertion} illustrates these requirements through plug insertion, contrasting a successful execution with two distinct failure modes: \emph{precision} failure and \emph{force} failure. In what follows, we characterize each failure mode in detail and investigate its connection to specific properties of flow-matching VLAs.

\subsection{Precision Failures}
\label{sec:visual}

Precision failures, such as the example in Figure~\ref{fig:plug-insertion-b}, occur when the policy cannot reliably produce the sub-millimeter corrections required during contact. We examine this difficulty in terms of action magnitude and the distribution of training samples across noise levels.

\vspace{-5pt}

\textbf{Action scale during contact.}
Contact-rich manipulation exhibits a strong distributional shift in action magnitude. In free space, action deltas are large and variable, but during contact they decrease to near zero (see Figure~\ref{fig:ee_profiles} in the Appendix for per-task delta profiles over time).
This low-delta regime must be reproduced precisely at the moment visual feedback is least informative, as contact occludes the precise alignment error and the scene appears nearly static during sub-millimeter corrections.

\begin{figure*}[!t]
\centering
\definecolor{hlfill}{HTML}{FFF5E6}
\definecolor{hldraw}{HTML}{B89968}
\definecolor{betacolor}{RGB}{195,162,222}
\definecolor{plncolor}{rgb}{0.957,0.835,0.647}

\begin{subfigure}[t]{0.37\textwidth}
\centering
\resizebox{\linewidth}{!}{%
\includegraphics{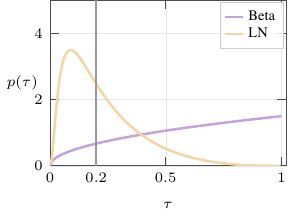}}
\caption{LN vs.\ Beta noise schedule}
\label{fig:arch_and_schedule}
\end{subfigure}
\hfill
\begin{subfigure}[t]{0.57\textwidth}
\centering
\resizebox{\linewidth}{!}{%
\includegraphics{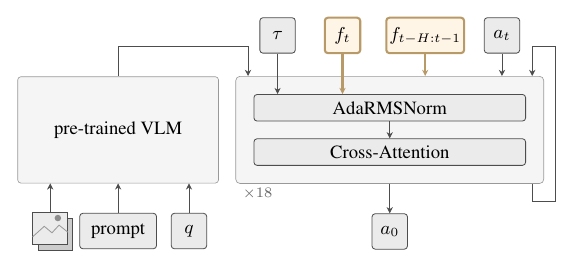}}
\caption{Force-augmented $\pi_{0.5}$ action expert}
\label{fig:arch_pi05}
\end{subfigure}

\caption{\textbf{FACT method overview.}
  \textbf{(a)} Logit-Normal (LN) assigns approximately $6{\times}$ more probability mass to the
  low-noise region ($\tau < 0.2$, gray line) than the
  Beta$(1.5,1.0)$ schedule.
  \textbf{(b)} Time-aware force injection uses the current reading $f_t$ to modulate per-layer sensitivity via AdaRMSNorm, while the force history $f_{t-H:t-1}$ is prepended as tokens for temporal context.}
  \vspace{-10pt}
\label{fig:method_overview}
\end{figure*}

\textbf{Low-noise sampling during post-training}.
During post-training, the Beta schedule used by $\pi_0$~\citep{black2024pi0}, $\pi_{0.5}$~\citep{black2025pi05}, and SmolVLA~\citep{shukor2025smolvla} allocates relatively few sampled timesteps to $\tau < 0.2$ (Figure~\ref{fig:arch_and_schedule}). This imbalance may hinder the acquisition of fine-grained corrections needed for reliable contact-rich behavior.

These observations motivate increasing low-noise sampling during post-training to improve the prediction of small corrective actions.

\vspace{-6pt}
\subsection{Force Failures}
\label{sec:force}
\vspace{-5pt}
Contact-rich tasks require policies to detect and react to forces, as illustrated in Figure~\ref{fig:plug-insertion-a}. However, even when force sensing is available, force-augmented VLAs struggle to use it effectively, leading to failures like the one in Figure~\ref{fig:plug-insertion-c}. Three properties motivate our force-injection design.

\textbf{Contact sparsity.}
Force signals are approximately zero over most free-space timesteps and become informative only during short contact intervals (see Figure~\ref{fig:ee_profiles} in the Appendix for per-task force profiles over time). This imbalance makes naive concatenation prone to underusing informative contact readings.

\textbf{Temporal structure.}
Force provides information at multiple timescales. The instantaneous measurement encodes the current interaction state, whereas recent history encodes the dynamics that led to it, including transients and cumulative force buildup. Omitting either timescale discards task-relevant contact information.

\textbf{Sensitivity modulation.}
The influence of force on the policy should depend on contact state. In free space, force readings should be effectively ignored, whereas at contact even small deviations should trigger corrective actions.

Collectively, these three properties motivate a force-injection mechanism that accounts for the imbalance of contact events, preserves the temporal structure of force signals, and uses force to modulate rather than simply augment the policy.

\section{Addressing Failures of VLA Policies in Contact-Rich Tasks}
\label{sec:method}
The failure modes identified in Section~\ref{sec:demistifying_failures} motivate two targeted interventions. To address precision failures, we replace the commonly used Beta noise schedule with the Logit-Normal (LN) schedule, which reallocates training signal toward the low-noise regime. To address force failures, we introduce a time-aware force injection mechanism that conditions the action generation module on contact state and a short force history. LN preserves the architecture, while force injection adds conditioning modules. Both require no additional data and add negligible inference-time overhead.

\subsection{Addressing Precision Failures: Logit-Normal Noise Schedule}

We adopt LN, an established sampling distribution for rectified flow training~\citep{esser2024scaling}, to increase exposure to low noise levels during post-training. Its distribution is given by:
\begin{equation}
    f_\mathcal{T}(\tau) = \frac{1}{s\sqrt{2\pi}}\frac{1}{\tau(1-\tau)}\exp{\left(-\frac{\left(\text{logit}(\tau)+m\right)^2}{2 s^2}\right)},
    \quad \text{logit}(\tau) = \ln \left(\frac{\tau}{1-\tau} \right).
\end{equation}
$\tau$ is sampled via the reparameterisation:
\begin{equation}
  \tau = \sigma(s \cdot z - m), \quad z \sim \mathcal{N}(0,1),
\end{equation}
where $\sigma$ is the sigmoid function. We use a location shift $m{=}1.5$ and scale $s{=}1$, shifting the distribution toward lower noise levels.
With this parameterization, LN allocates $6{\times}$ more sampling probability to $\tau < 0.2$ than the Beta schedule as illustrated in Figure~\ref{fig:arch_and_schedule}. This concentrates post-training on reconstructing actions from small perturbations, with the aim of improving the precise corrective actions required during contact.
Critically, LN requires no changes to the model
architecture, no additional training data, and no extra parameters, making it
compatible with flow-matching objectives. Performance is tested on two VLA families.

\subsection{Addressing Force Failures: Time-Aware Force Injection}

Our force injection approach, illustrated in Figure~\ref{fig:arch_pi05}, is built around three design choices, each targeting one of the force properties identified in Section~\ref{sec:force}.

\begin{figure*}[!t]
\centering

\newcommand{\axshifttask}{15pt}
\newcommand{\imgH}{48pt}
\newcommand{\gaptask}{2pt}

\resizebox{\linewidth}{!}{%
\includegraphics{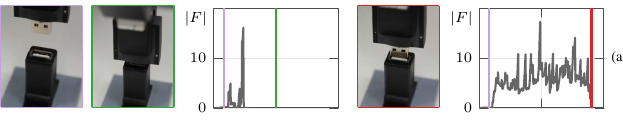}}

\vspace{1pt}

\resizebox{\linewidth}{!}{%
\includegraphics{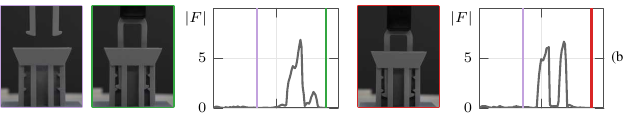}}

\caption{\textbf{VLA failures on two contact-rich tasks.}
  Each row shows an approach frame (purple) followed by a success (green) and failure (red) frame with the corresponding force profiles. Vertical lines denote the approach event and the times at which success or failure is registered. In USB insertion, failure is caused by sub-millimeter pose misalignment. In key insertion, failure is caused by missing the hard-stop and continuing to push past the seating point. The key is shown unoccluded for clarity although the lock is fully occluded during execution.}
\vspace{-10pt}
\label{fig:task_force_profiles}
\phantomcaption\label{fig:task_force_profiles-a}
\phantomcaption\label{fig:task_force_profiles-b}
\end{figure*}

\paragraph{Contact state.} Targeting sensitivity modulation, we inject the current force reading into the normalization layers of the action generation module to adapt the policy's responsiveness to the contact state. The most recent sensor window $\mathbf{f}_t \in \mathbb{R}^{H_w \times 6}$ is mean-pooled to a
6-dimensional summary $\bar{f}_t$, passed through a two-layer MLP, and
projected to a per-layer scale modulation:
\begin{equation}
  \Delta\gamma(\bar{f}_t) = W_\gamma\,\phi_{\mathrm{force}}(\bar{f}_t) \in \mathbb{R}^{d}.
\end{equation}
This correction is added to the AdaRMS scale at every layer:
\begin{equation}
  h_l = \bigl(\gamma_l(\tau) + \Delta\gamma(\bar{f}_t)\bigr)\cdot
        \mathrm{RMSNorm}(h_{l-1}) + \beta_l(\tau) + h_{l-1}\,g_l(\tau).
\end{equation}
$\Delta\gamma$ is shared across all layers and modulates the sensitivity of every layer in a single forward pass.

\paragraph{Contact history.}
Targeting temporal structure, the preceding $H{=}30$ steps (2\,s) of F/T readings $\{\mathbf{f}_{t-H},\ldots,\mathbf{f}_{t-1}\}$ are each encoded independently by a shared causal encoder~\citep{bai2018tcn}. The resulting tokens are prepended to the input of the action generation module, enabling the policy to reason over the trajectory of contact.

\vspace{-6pt}
\paragraph{Contact gating.} Targeting contact sparsity, a gradient gate with threshold $\delta$ blocks gradient through all force encoding components for steps where no contact is detected, preventing the force encoders from fitting to uninformative near-zero readings.

Together, the LN noise schedule and the time-aware force injection form \method, Force-Aware Contact-rich manipulation via Timestep Modulation. LN increases low-noise sampling during post-training to improve the prediction of small corrective actions, while time-aware force injection integrates contact state and force history to guide action generation during contact. In Section~\ref{sec:results}, we validate the generality of LN when applied to existing force-augmented baselines~\citep{tavla2025,yu2025forcevla} and show that \method\ transfers across flow-matching backbones.

\vspace{-6pt}
\section{Experimental Setup}
\label{sec:setup}
\vspace{-6pt}
\begin{table*}[!t]
  \centering
  \caption{\method\ outperforms all baselines across five contact-rich tasks, with consistent gains on both precision- and force-critical tasks. Tasks from left to right: plug insertion, USB insertion, button push, board erasing, and key insertion.}
  \label{tab:results}
  \setlength{\tabcolsep}{3pt}
  \setlength{\aboverulesep}{0pt}
  \setlength{\belowrulesep}{0pt}
  \resizebox{\textwidth}{!}{%
  \scriptsize
  \begin{tabular}{l | rc | rc | rc | rc | rc | c}
    \toprule
    &
    \multicolumn{2}{c|}{\includegraphics[height=21pt]{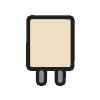}} &
    \multicolumn{2}{c|}{\includegraphics[height=21pt]{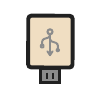}} &
    \multicolumn{2}{c|}{\includegraphics[height=21pt]{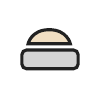}} &
    \multicolumn{2}{c|}{\includegraphics[height=21pt]{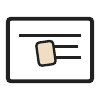}} &
    \multicolumn{2}{c|}{\includegraphics[height=21pt]{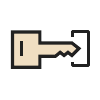}} &
    \textbf{All} \\
    \textbf{Method}
      & $\uparrow$SR{\tiny[\%]} & $\downarrow p$
      & $\uparrow$SR{\tiny[\%]} & $\downarrow p$
      & $\uparrow$SR{\tiny[\%]} & $\downarrow p$
      & $\uparrow$SR{\tiny[\%]} & $\downarrow p$
      & $\uparrow$SR{\tiny[\%]} & $\downarrow p$
      & $\uparrow$SR{\tiny[\%]} \\
    \midrule
    $\pi_{0.5}$
      & 30.0 & --
      & 37.5 & --
      & 12.5 & --
      & \textbf{100.0} & --
      & 15.0 & --
      & 39.0 \\
    $\pi_{0.5}$ + LN
      & 50.0 & $.055$
      & 47.5 & $.249$
      & 57.5 & $<.001$
      & 87.5 & $1.00$
      & 37.5 & $.020$
      & 56.0 \\
    \method
      & \textbf{57.5} & $.012$
      & \textbf{47.5} & $.249$
      & \textbf{75.0} & $<.001$
      & 90.0           & $1.00$
      & \textbf{60.0} & $<.001$
      & \textbf{66.0} \\
    \midrule
    ForceVLA$_{\pi_{0.5}}$
      & 32.5 & $.500$
      & 37.5 & $.591$
      & 12.5 & $.631$
      & 77.5 & $1.00$
      & 42.5 & $.006$
      & 40.5 \\
    TA-VLA$_{\pi_{0.5}}$
      & 30.0 & $.596$
      & 25.0 & $.926$
      & 20.0 & $.273$
      & 97.5 & $1.00$
      & 15.0 & $.622$
      & 37.5 \\
    \bottomrule
  \end{tabular}%
  }
\end{table*}

We describe the tasks, data collection, baselines, and evaluation protocol used in our experiments.

\paragraph{Tasks}

We evaluate on five contact-rich tasks spanning both failure modes, with full descriptions and task illustrations provided in Appendix~\ref{app:tasks}. Plug insertion and USB insertion are precision-critical, demanding sub-millimeter corrections under partial occlusion. Key insertion, button push, and board erasing are force-critical. Key insertion is fully occluded and requires recognizing the hard-stop force signature as the key seats. Button push requires probing until a force threshold is reached, which locks the button in place. Board erasing requires maintaining consistent contact throughout the task. Representative episodes for USB and key insertion are included in Figure~\ref{fig:task_force_profiles}.

\paragraph{Data Collection and Training}

All experiments are conducted on a Franka Research 3 arm equipped with a wrist-mounted Bota SensONE F/T sensor (see Appendix~\ref{app:hardware} for hardware details). For each task, we collect 100 teleoperated demonstrations using a Haply Inverse 3 haptic device that provides force feedback to the operator. At the start of each episode, the target position is sampled uniformly over a $32{\times}20$\,cm surface to encourage robustness, and the robot's home position is sampled uniformly within a $5$\,cm cube. Visual and proprioceptive observations are recorded at 15\,Hz and synchronized with F/T readings recorded at 400\,Hz. All methods, including baselines, are fine-tuned from the pre-trained $\pi_{0.5}$ checkpoint using LoRA for 20{,}000 steps on the same demonstrations.

\paragraph{Baselines}
We compare \method against the standard $\pi_{0.5}$ policy and two force-conditioned baselines, ForceVLA~\citep{yu2025forcevla} and TA-VLA~\citep{tavla2025}. Although ForceVLA and TA-VLA were originally introduced using the $\pi_{0}$ architecture, we re-implement both methods on top of $\pi_{0.5}$ to enable a controlled comparison under a shared policy backbone. We additionally report $\pi_{0}$ backbone results in Table~\ref{tab:pi0} to enable direct comparison with the originally proposed baseline configurations. For a detailed comparison of \method against baseline architectures please refer to Appendix~\ref{app:baselines}.
\vspace{-6pt}
\paragraph{Metrics and Evaluation Protocol}

Each method is evaluated on 40 independent rollouts per task, with target and home poses sampled from the same distributions used during training. Across all conditions, this yields nearly \emph{2{,}500} evaluation rollouts in total. A rollout is considered successful if the task is completed within a 60\,s timeout. We report success rate (SR\,\%) as the fraction of successful rollouts and express differences in percentage points (pp). We report one-sided Fisher's exact tests for improvement over the $\pi_{0.5}$ baseline (see Appendix~\ref{app:statistics} for details).

\vspace{-6pt}
\section{Results and Discussion}
\label{sec:results}
\vspace{-6pt}

\begin{table*}[!t]
  \centering
  \setlength{\aboverulesep}{0pt}
  \setlength{\belowrulesep}{0pt}
  \setlength{\tabcolsep}{3pt}
  \begin{minipage}[t]{0.31\linewidth}
    \vspace{0pt}
    \centering
    \captionof{table}{LN improves the performance of baselines
      with different force architectures built on the same $\pi_{0.5}$ backbone.}
    \label{tab:baselines_pi05}
    \resizebox{\linewidth}{!}{\small
    \begin{tabular}{l | c | c | c}
      \toprule
      &
      \includegraphics[height=22pt]{images/icons-pdf/plug-insertion} &
      \includegraphics[height=22pt]{images/icons-pdf/key-insertion} &
      \includegraphics[height=22pt]{images/icons-pdf/button-pushing} \\
      \textbf{Method}
        & $\uparrow$SR{\tiny[\%]}
        & $\uparrow$SR{\tiny[\%]}
        & $\uparrow$SR{\tiny[\%]} \\
      \midrule
      TA-VLA                & 30.0 & 15.0 & 20.0 \\
      \hspace{4pt}+ LN      & \textbf{47.5} & \textbf{32.5} & \textbf{42.5} \\
      \midrule
      ForceVLA              & 32.5 & \textbf{42.5} & 12.5 \\
      \hspace{4pt}+ LN      & \textbf{57.5} & 30.0 & \textbf{30.0} \\
      \bottomrule
    \end{tabular}}
  \end{minipage}
  \hspace{4pt}
  \begin{minipage}[t]{0.31\linewidth}
    \vspace{0pt}
    \centering
    \captionof{table}{Ablating force components from \method\ shows all are necessary, with the largest drops on force-critical tasks.}
    \label{tab:ablations}
    \resizebox{\linewidth}{!}{\small
    \begin{tabular}{l | c | c | c}
      \toprule
      &
      \includegraphics[height=20pt]{images/icons-pdf/plug-insertion} &
      \includegraphics[height=20pt]{images/icons-pdf/key-insertion} &
      \includegraphics[height=20pt]{images/icons-pdf/button-pushing} \\
      \textbf{Method}
        & SR{\tiny[\%]}
        & SR{\tiny[\%]}
        & SR{\tiny[\%]} \\
      \midrule
      \method               & 57.5 & \textbf{60.0} & \textbf{75.0} \\
      \midrule
      w/o grad th           & 42.5 & 27.5 & 57.5 \\
      w/o curr.\ read       & \textbf{60.0} & 35.0 & 50.0 \\
      w/o history           & 30.0 & 20.0 & 12.5 \\
      \bottomrule
    \end{tabular}}
  \end{minipage}
  \hspace{4pt}
  \begin{minipage}[t]{0.33\linewidth}
    \vspace{0pt}
    \centering
    \captionof{table}{\method applied to the $\pi_0$ backbone alongside prior
      baselines, showing that gains transfer across backbone versions.}
    \label{tab:pi0}
    \resizebox{\linewidth}{!}{\small
    \begin{tabular}{l | c | c | c}
      \toprule
      &
      \includegraphics[height=22pt]{images/icons-pdf/plug-insertion} &
      \includegraphics[height=22pt]{images/icons-pdf/key-insertion} &
      \includegraphics[height=22pt]{images/icons-pdf/button-pushing} \\
      \textbf{Method}
        & $\uparrow$SR{\tiny[\%]}
        & $\uparrow$SR{\tiny[\%]}
        & $\uparrow$SR{\tiny[\%]} \\
      \midrule
      $\pi_0$               & 32.5 & 45.0 & 47.5 \\
      \methodpizero     & \textbf{70.0} & \textbf{50.0} & \textbf{60.0} \\
      \midrule
      ForceVLA$_{\pi_0}$    & 55.0 & 42.5 & 42.5 \\
      TA-VLA$_{\pi_0}$      & 30.0 & 27.5 & 27.5 \\
      \bottomrule
    \end{tabular}}
  \end{minipage}
\end{table*}

We structure the results around six experimental questions, examining the two failure modes and the components designed to address them.

\paragraph{Does LN improve flow-based VLAs performance on precision-critical tasks?}
\label{sec:q1}

Replacing the Beta schedule with LN improves plug insertion by 20\,pp ($p{=}.055$) and USB insertion by 10\,pp. LN gains are also observed on force-critical tasks. Button increases by $+$45\,pp with $p{<}.001$ and key insertion $+$22.5\,pp with $p{=}.020$. These gains require no additional data or parameters.

\paragraph{Does time-aware force injection improve performance on force-critical tasks?}
\label{sec:q2}
Adding time-aware force injection on top of $\pi_{0.5}{+}$LN (\method row in Table~\ref{tab:results}) yields $+$17.5\,pp on button push and $+$22.5\,pp on key insertion, confirming that force reasoning at contact is essential for tasks where visual observation is ambiguous, with $p{<}.001$ for both against $\pi_{0.5}$. On precision-critical tasks, the added force signal provides no statistically significant benefit. Board erasing stands out as an outlier among force-critical tasks, with near-perfect success rates across all methods. Sustained surface contact is largely handled by the compliant operational-space controller (Appendix~\ref{app:controller}), which maintains consistent contact force without requiring explicit force reasoning from the policy. The sparse failures observed are due to visual misalignment causing the eraser to cover only part of the mark rather than failing to maintain contact with the surface.
\paragraph{How does each force component address the identified contact properties?}
\label{sec:q4}

Table~\ref{tab:ablations} ablates each force component. Force history matters most, with button push dropping by $-$62.5\,pp and key insertion by $-$40\,pp, confirming that temporal integration is essential for force-critical tasks. The instantaneous reading plays a smaller role, except on key insertion ($-$25\,pp) where the hard-stop peak provides a completion cue that history alone cannot supply. The gradient threshold proves necessary on force-critical tasks, where contact events carry the decisive completion signal. To verify these gains reflect genuine force exploitation rather than a training regularization, we replace F/T readings with Gaussian noise during training and evaluation. Full results are in Appendix~\ref{app:noise_ablation}.

\vspace{-6pt}
\paragraph{Does LN recover baseline performance on precision-critical tasks?}
\label{sec:q4pln}

Table~\ref{tab:baselines_pi05} shows that adding LN to either baseline yields consistent gains. This supports that LN can be used as a drop-in fix for other flow-based VLAs without requiring any architectural changes.

\vspace{-6pt}
\paragraph{Do LN and time-aware force injection transfer to other VLA backbones like $\pi_0$?}
\label{sec:q5}

Table~\ref{tab:pi0} evaluates \methodpizero alongside the
baselines on the $\pi_0$ backbone. \methodpizero achieves 70.0\,\% on plug insertion,
50.0\,\% on key insertion, and 60.0\,\% on button push, outperforming both
\forcevla and \tavla. This shows that \method transfers across backbone versions. Comparing \methodpizero against \methodpizeroft, $\pi_{0.5}$ is higher on force-critical tasks by over $+$10\,pp, while $\pi_0$ is higher on the precision-critical task by $+$12.5\,pp. The $\pi_{0.5}$ advantage on force-critical tasks may reflect its more explicit and pretrained timestep conditioning, which could make the force injection more effective at modulating the action expert's sensitivity. The $\pi_0$ advantage on precision-critical tasks may come from differences in pretraining or state representation that improve fine corrections. Beyond the $\pi$ series, GR00T N1 plug insertion improves from 40\% to 65\% with LN alone. On key insertion, \method improves success from 15\% to 35\%, while LN alone achieves 12.5\%. These results extend the evidence to a second VLA family.

\paragraph{How does \method perform on standard manipulation tasks?}
On a standard bowl-stacking task, FACT achieves a 75\% success rate, compared with 70\% for $\pi_{0.5}$. Because this task is simpler than the contact-rich tasks of this study, we fine-tune both methods using 15 demonstrations for 5K steps, rather than 100 demonstrations for 20K steps. These results suggest that FACT maintains performance on this standard manipulation task and offers limited additional benefit. Broader evaluation is needed to determine how LN during post-training affects tasks that do not require fine-grained corrections.

Overall, \methodpizeroft achieves 66.0\,\% success, outperforming ForceVLA$_{\pi_{0.5}}$ (40.5\,\%) and TA-VLA$_{\pi_{0.5}}$ (37.5\,\%) with consistent gains on both precision- and force-critical tasks, suggesting that addressing each failure mode independently leads to more consistent gains across contact-rich tasks.

\vspace{-6pt}
\section{Conclusion}
\label{sec:conclusion}
\vspace{-6pt}
We present \method, a method that addresses two distinct failure modes of flow-based VLAs on contact-rich tasks. We identify precision failures and force failures, and propose targeted solutions for each. The LN noise schedule reallocates training samples to the low-noise denoising regime, yielding consistent gains on precision-critical tasks without additional parameters or data. The time-aware force injection mechanism captures the sparsity and temporal structure of force signals, yielding large gains on force-critical tasks where visual feedback is insufficient. The two components are complementary. LN improves the prediction of small corrective actions, while structured force conditioning provides complementary gains on tasks requiring force feedback. Results on the $\pi$ series and GR00T N1 demonstrate transfer across two VLA families, suggesting they address general limitations of flow-based VLAs.

\paragraph{Limitations.}
Our evaluation is conducted on a single robot platform with a fixed wrist-mounted F/T sensor, leaving open how well the force injection mechanism generalizes to robots with different kinematic structures or sensor placements. LN is specific to flow-matching action heads and does not directly apply to autoregressive or diffusion policies. Time-aware force injection is designed for action heads with transformer layers modulated via RMS scaling and would require adaptation for different architectures. Finally, broader evaluation on unseen objects and additional standard manipulation tasks, together with comparisons of force conditioning methods under a shared architecture, would strengthen the generality of our conclusions.

\subsubsection*{Acknowledgements}
This work was supported in part by Agile Robotics. Carlota Parés-Morlans is supported by a graduate fellowship from Knight-Hennessy Scholars at Stanford University. Nils Kuhn is supported by scholarships from the Friedrich Ebert Foundation and the German Academic Exchange Service (DAAD). Alberta Longhini is supported by a Wallenberg--Bienenstock Postdoctoral Fellowship.
We thank Hila Chefer for helpful discussions, and Michelle Yi for discussions on environment setup, data collection, and camera footage for figures.

\clearpage
\bibliography{main}

\clearpage
\appendix
\renewcommand{\thefigure}{A.\arabic{figure}}
\setcounter{figure}{0}
\renewcommand{\thetable}{A.\arabic{table}}
\setcounter{table}{0}

\section{Task Descriptions}
\label{app:tasks}

\begin{figure*}[!b]
  \centering
  \setlength{\tabcolsep}{1pt}
  \begin{tabular}{ccccc}
    \includegraphics[height=22pt]{images/icons-pdf/plug-insertion} &
    \includegraphics[height=22pt]{images/icons-pdf/usb-insertion} &
    \includegraphics[height=22pt]{images/icons-pdf/board-erasing} &
    \includegraphics[height=22pt]{images/icons-pdf/button-pushing} &
    \includegraphics[height=22pt]{images/icons-pdf/key-insertion} \\[1pt]
    \begin{subfigure}[t]{0.19\linewidth}\centering
      \includegraphics[width=\linewidth]{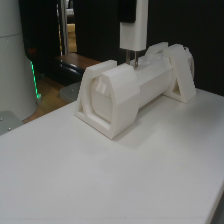}\\[0pt]
      \includegraphics[width=\linewidth]{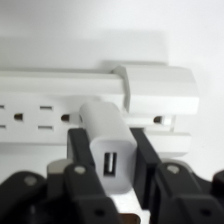}
      \phantomcaption\label{fig:task_views_grid-a}
    \end{subfigure} &
    \begin{subfigure}[t]{0.19\linewidth}\centering
      \includegraphics[width=\linewidth]{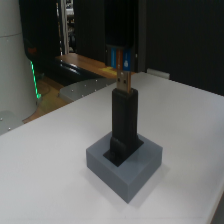}\\[0pt]
      \includegraphics[width=\linewidth]{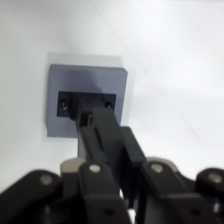}
      \phantomcaption\label{fig:task_views_grid-b}
    \end{subfigure} &
    \begin{subfigure}[t]{0.19\linewidth}\centering
      \includegraphics[width=\linewidth]{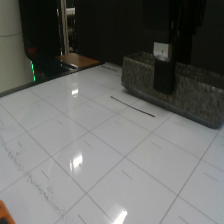}\\[0pt]
      \includegraphics[width=\linewidth]{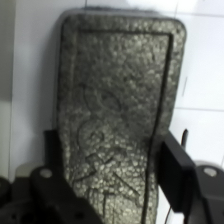}
      \phantomcaption\label{fig:task_views_grid-c}
    \end{subfigure} &
    \begin{subfigure}[t]{0.19\linewidth}\centering
      \includegraphics[width=\linewidth]{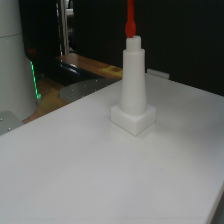}\\[0pt]
      \includegraphics[width=\linewidth]{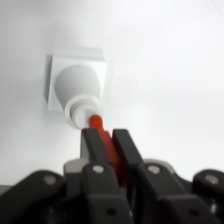}
      \phantomcaption\label{fig:task_views_grid-d}
    \end{subfigure} &
    \begin{subfigure}[t]{0.19\linewidth}\centering
      \includegraphics[width=\linewidth]{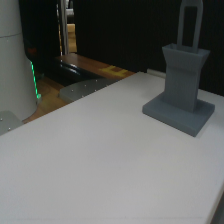}\\[0pt]
      \includegraphics[width=\linewidth]{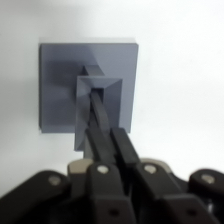}
      \phantomcaption\label{fig:task_views_grid-e}
    \end{subfigure} \\
    \small (a) & \small (b) & \small (c) & \small (d) & \small (e) \\
  \end{tabular}
  \caption{External (top) and wrist-mounted (bottom) camera views for each task at a representative contact moment. From left to right: (\subref{fig:task_views_grid-a}) plug insertion, (\subref{fig:task_views_grid-b}) USB insertion, (\subref{fig:task_views_grid-c}) board erasing, (\subref{fig:task_views_grid-d}) button push, (\subref{fig:task_views_grid-e}) key insertion.}
  \label{fig:task_views_grid}
\end{figure*}

We evaluate \method on the five contact-rich manipulation tasks illustrated in Figure~\ref{fig:task_views_grid}.

\begin{figure*}[!b]
\centering

\newcommand{\axshift}{15pt}
\newcommand{\imgstep}{54pt}

\newcommand{\rcolwidth}{0.64\linewidth}
\newcommand{\lcolwidth}{\dimexpr\linewidth-\rcolwidth-6pt\relax}

\noindent
\adjustbox{valign=t}{\begin{minipage}{\lcolwidth}%
  \includegraphics[width=\linewidth]{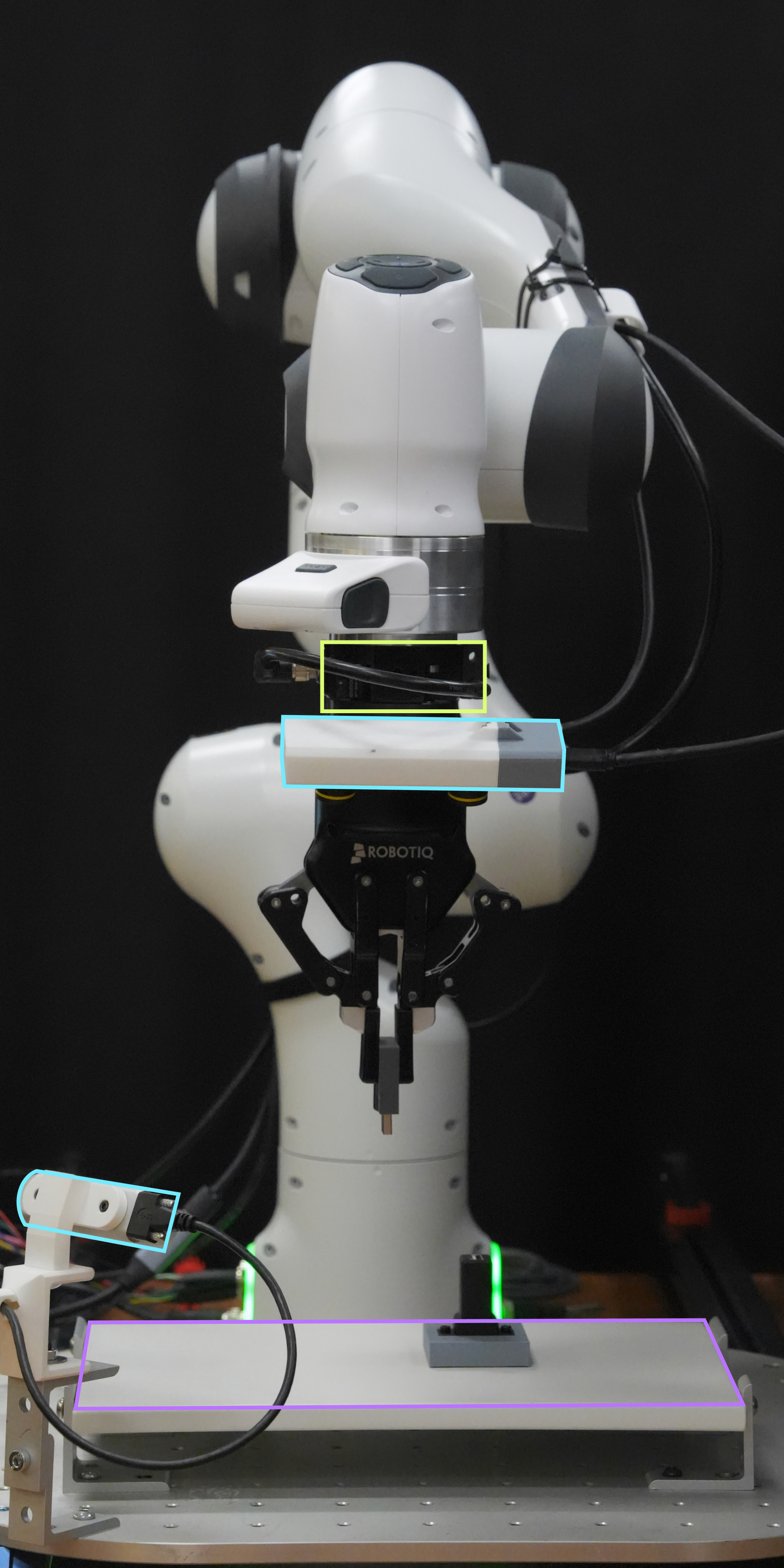}%
\end{minipage}}%
\hskip 6pt%
\adjustbox{valign=t}{\begin{minipage}{\rcolwidth}%
\captionsetup[subfigure]{aboveskip=1pt, belowskip=0pt}

\begin{subfigure}[b]{\linewidth}
\centering
\resizebox{\linewidth}{!}{%
\includegraphics{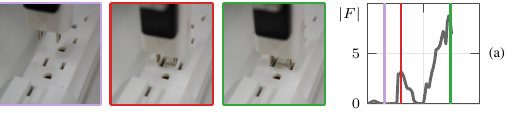}}
\phantomcaption
\label{fig:task_force_profiles_backup-a}
\end{subfigure}\vspace{-12pt}

\begin{subfigure}[b]{\linewidth}
\centering
\resizebox{\linewidth}{!}{%
\includegraphics{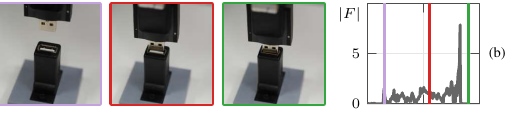}}
\phantomcaption
\label{fig:task_force_profiles_backup-b}
\end{subfigure}\vspace{-12pt}

\begin{subfigure}[b]{\linewidth}
\centering
\resizebox{\linewidth}{!}{%
\includegraphics{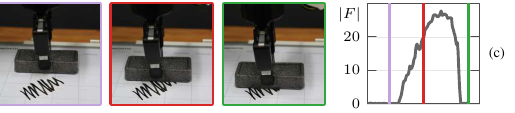}}
\phantomcaption
\label{fig:task_force_profiles_backup-c}
\end{subfigure}\vspace{-12pt}

\begin{subfigure}[b]{\linewidth}
\centering
\resizebox{\linewidth}{!}{%
\includegraphics{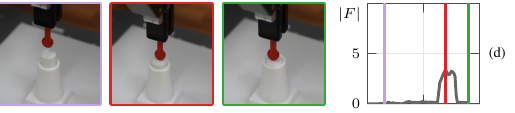}}
\phantomcaption
\label{fig:task_force_profiles_backup-d}
\end{subfigure}\vspace{-12pt}

\begin{subfigure}[b]{\linewidth}
\centering
\resizebox{\linewidth}{!}{%
\includegraphics{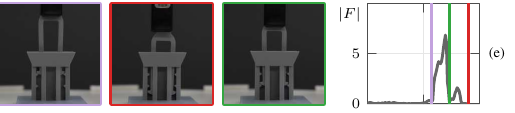}}
\phantomcaption
\label{fig:task_force_profiles_backup-e}
\end{subfigure}

\end{minipage}}

\caption{\textbf{Environment state and force profile for five tasks.}
  Columns show three representative episode moments followed by the full force profile. Purple marks an approaching step before contact, red marks the dominant failure point, and green marks success. (a) plug insertion, (b) USB insertion, (c) board erasing, (d) button push, and (e) key insertion. Key insertion images show the lock with an open back for visualization purposes. The left column shows the robot hardware setup: the {\setlength{\fboxrule}{1.2pt}\setlength{\fboxsep}{1pt}\fcolorbox{yellow}{white}{yellow}} box highlights the Bota force/torque sensor mounted between the robot wrist and the Robotiq gripper, {\setlength{\fboxrule}{1.2pt}\setlength{\fboxsep}{1pt}\fcolorbox{cyan}{white}{blue}} boxes mark the wrist-mounted camera and the external camera, and the {\setlength{\fboxrule}{1.2pt}\setlength{\fboxsep}{1pt}\fcolorbox{violet}{white}{purple}} box outlines the ferromagnetic board on which task objects are repositioned between trials.}
\label{fig:task_force_profiles_all}
\end{figure*}

\paragraph{Plug insertion} Illustrated in Figure~\ref{fig:task_views_grid-a}, a two-pin power plug must be inserted into an extension socket. Once the plug approaches the socket,
the connector body occludes the holes, removing visual confirmation of alignment from the wrist camera.
Seating the plug requires sustained force to overcome the friction fit, as shown in the force profile of Figure~\ref{fig:task_force_profiles_backup-a}.

\paragraph{USB insertion} Illustrated in Figure~\ref{fig:task_views_grid-b}, a USB-A connector must be inserted into a port. During insertion, the connector body occludes the port from the wrist-mounted camera, making visual alignment unreliable. Seating the connector requires sustained force to overcome the friction fit, as shown in the force profile of Figure~\ref{fig:task_force_profiles_backup-b}.

\paragraph{Board eraser} Illustrated in Figure~\ref{fig:task_views_grid-c}, a board eraser must be moved across a whiteboard surface to erase a marked region. The task requires maintaining consistent contact throughout the stroke. Too little force leaves residue, while too much risks skipping or damaging the surface. Force feedback is the primary signal for regulating contact quality, as shown in Figure~\ref{fig:task_force_profiles_backup-c}.

\paragraph{Button push} Illustrated in Figure~\ref{fig:task_views_grid-d}, a button must be pressed by probing until a force threshold is reached, which in turn locks it in place. Demonstrations were collected such that pressing with maximum force is not a viable strategy. The contact force shown in Figure~\ref{fig:task_force_profiles_backup-d} provides the primary signal for determining when sufficient pressure has been applied, as vision alone is unreliable for detecting the locking event.

\paragraph{Key insertion} Illustrated in Figure~\ref{fig:task_views_grid-e}, a key must be inserted into a lock and seated at the correct depth, neither undershooting nor overshooting. To force reliance on force signals, demonstrations were collected using visually identical keys of different lengths, creating deliberate visual ambiguity. The force profile shown in Figure~\ref{fig:task_force_profiles_backup-e} provides the primary signal for detecting when the correct seating depth has been reached.

\section{Hardware Setup}
\label{app:environment}
\label{app:hardware}

All experiments are conducted on a single \textbf{Franka Research 3} arm equipped with a
wrist-mounted \textbf{Bota SensONE} 6-axis force/torque (F/T) sensor. The full setup is illustrated in Figure~\ref{fig:task_force_profiles_all}.
The robot is controlled via Cartesian impedance control at 1\,kHz. Futher details on the controller design are provided in Appendix~\ref{app:controller}. Visual observations are provided by an external Realsense D435 camera and a wrist-mounted ZED Mini camera, both operating at 15\,Hz. F/T data is acquired at 400\,Hz and pre-processed with a one-euro filter \cite{casiez20121}.
Each control step therefore receives a window of $H_w{=}27$ raw sensor readings
($\lceil 400/15 \rceil{=}27$).

All task objects were mounted on 3D-printed bases with embedded magnets, secured to a ferromagnetic board to prevent displacement during execution while allowing repositioning between trials. The plug insertion target was a commercial power strip extension held in a 3D-printed magnetic mount. The USB insertion used the connector from the NIST Robotic Assembly Board in a 3D-printed magnetic mount. The button push and key lock mechanisms were fully 3D-printed. For board erasing, the board was fixed and the drawn mark to be erased was varied between trials. A full bill of materials and 3D-printable STL files will be released alongside the code. Across all tasks, object positions were sampled uniformly within a $32{\times}20$\,cm workspace.

\begin{table}[!t]
  \centering
  \setlength{\aboverulesep}{0pt}
  \setlength{\belowrulesep}{0pt}
  \setlength{\tabcolsep}{3pt}
  \caption{Noise substitution ablation replacing the measured F/T stream with i.i.d.\ Gaussian noise, preserving input dimensionality while removing force information specific to the task. \method shows significant drops on key insertion and button push ($p{<}.001$), but not on plug insertion ($p{=}.090$), supporting its use of force information on the former two tasks. In contrast, \forcevla shows a significant drop only on key insertion, and \tavla shows no statistically significant drop, suggesting their architectures do not reliably extract force information.}
  \label{tab:force_noise}
  \vspace{4pt}
  \begin{tabularx}{0.8\linewidth}{>{\raggedright\arraybackslash}X | r c | r c | r c}
    \toprule
    &
    \multicolumn{2}{c|}{\includegraphics[height=22pt]{images/icons-pdf/plug-insertion}} &
    \multicolumn{2}{c|}{\includegraphics[height=22pt]{images/icons-pdf/key-insertion}} &
    \multicolumn{2}{c}{\includegraphics[height=22pt]{images/icons-pdf/button-pushing}} \\
    \textbf{Method}
      & $\uparrow$SR{\tiny[\%]} & $\downarrow p$
      & $\uparrow$SR{\tiny[\%]} & $\downarrow p$
      & $\uparrow$SR{\tiny[\%]} & $\downarrow p$ \\
    \midrule
    TA-VLA
      & \textbf{30.0} & $.306$
      & 15.0 & $.919$
      & \textbf{20.0} & $.273$ \\
    TA-VLA force $\to$ noise
      & 22.5 & --
      & \textbf{25.0} & --
      & 12.5 & -- \\
    \midrule
    ForceVLA
      & 32.5 & $.824$
      & \textbf{42.5} & $.006$
      & 12.5 & $.826$ \\
    ForceVLA force $\to$ noise
      & \textbf{40.0} & --
      & 15.0 & --
      & \textbf{17.5} & -- \\
    \midrule
    \method
      & \textbf{57.5} & $.090$
      & \textbf{60.0} & $<.001$
      & \textbf{75.0} & $<.001$ \\
    \method force $\to$ noise
      & 40.0 & --
      & 5.0 & --
      & 17.5 & -- \\
    \bottomrule
  \end{tabularx}
\end{table}

\section{Controller Design}
\label{app:controller}
We use a two-rate control architecture that decouples 15\,Hz policy updates from a 1\,kHz joint-torque control loop, enabling compliant behavior during contact via operational-space control ~\citep{khatib1987, lee2020makingsense}. Let $\mathbf{x} \in \mathbb{R}^3$ denote end-effector position and $R \in \mathrm{SO}(3)$ its orientation. The policy outputs Cartesian end-effector displacement commands $\Delta\mathbf{x} \in \mathbb{R}^3$ and angle displacements $\Delta\alpha \in \mathbb{R}^3$. The controller framework comprises three stages of trajectory generation, impedance control, and operational-space torque computation.
\vspace{-5pt}
\paragraph{Trajectory generation.}
At each policy step, the trajectory generator computes a desired end-effector
pose $\mathbf{p}_{\mathrm{des}}$ from the current pose $\mathbf{p}_t$ and the
commanded displacements $\Delta\mathbf{x}$ and $\Delta\alpha$.
It then interpolates a smooth trajectory
$\xi_t = \{\mathbf{p}_k, \mathbf{v}_k, \mathbf{a}_k\}_{k=t}^{t+T}$
of position, velocity, and acceleration at 1\,kHz, bridging the gap
between the low-bandwidth policy and the high-bandwidth controller.
\vspace{-5pt}
\paragraph{Impedance control.}
A Cartesian impedance PD controller tracks the interpolated trajectory and
computes a task-space acceleration command:
\begin{equation}
  \mathbf{a}_u =
     \mathbf{a}_{\mathrm{des}}
     - \mathbf{k}_p(\mathbf{x} - \mathbf{x}_{\mathrm{des}})
     - \mathbf{k}_v(\mathbf{v} - \mathbf{v}_{\mathrm{des}}),
\end{equation}
where $\mathbf{k}_p$ and $\mathbf{k}_v$ are manually tuned stiffness and damping gains.
Compliance during contact makes the robot safer and allows it to slide
along surfaces under uncertainty, which is beneficial for contact-rich tasks.
\vspace{-5pt}
\paragraph{Operational-space torque computation.}
Using the known kinematic and dynamic model of the robot, we compute joint torques from Cartesian-space accelerations via the dynamically consistent operational-space formulation~\citep{khatib1987}. The task-space acceleration $\mathbf{a}_u$ is first mapped to a task-space force via the operational-space inertia matrix $\Lambda(\mathbf{q})$,
\begin{equation}
  \mathbf{f} = \Lambda(\mathbf{q})\,\mathbf{a}_u,
\end{equation}
and then to joint torques:
\begin{equation}
  \boldsymbol{\tau}_u = J^\top(\mathbf{q})\,\mathbf{f},
\end{equation}
where $J(\mathbf{q})$ is the Jacobian at joint configuration $\mathbf{q}$ and
$\Lambda(\mathbf{q})$ is the joint-space inertia matrix.

\vspace{-5pt}
\section{Noise Ablation}
\label{app:noise_ablation}

To test whether the F/T input provides task-relevant haptic information rather than merely acting as an auxiliary input channel that regularises learning, we perform the ablation study shown in Table~\ref{tab:force_noise}. Specifically, we replace the measured F/T stream with i.i.d.\ zero-mean Gaussian noise with unit standard deviation. This preserves the input dimensionality and network pathway associated with F/T while removing task-dependent force cues. To match inference-time conditions, policies are trained from scratch with Gaussian noise in place of the measured F/T readings.

Replacing F/T with Gaussian noise yields no statistically significant change on plug insertion. In contrast, \method shows significant drops on key insertion and button push ($p<.001$), supporting its use of force information on these tasks. \forcevla also benefits from real F/T on key insertion, where success drops from 42.5\,\% to 15.0\,\% under noise replacement. \tavla shows no statistically significant differences on key insertion or button push.

\section{Baseline Architecture Comparison}
\label{app:baselines}

\begin{table}[!ht]
  \centering
  \setlength{\aboverulesep}{0pt}
  \setlength{\belowrulesep}{0pt}
  \setlength{\tabcolsep}{3pt}
  \caption{Comparison of force-augmented VLA architectures.
    All methods use the $\pi_{0.5}$ backbone and are fine-tuned with LoRA. Parameter counts refer to added modules only. AE = action expert.}
  \label{tab:arch_comparison}
  \vspace{4pt}
  \resizebox{\columnwidth}{!}{%
  \scriptsize
  \begin{tabular}{l|c|c|c}
    \toprule
    & \textbf{\method} & \textbf{\forcevla} & \textbf{\tavla} \\
    \midrule
    Force encoder       & Shared causal TCN + 2-layer MLP  & Linear proj.\ + LIMoE  & 2-layer MLP \\
    Injection point     & AE (AdaRMS + tokens)             & VLM/AE bridge (MoE)    & AE (1 token) \\
    Temporal encoding   & $H{=}30$ windows of 27 steps (${\approx}2$\,s) & None & 10 frames (${\approx}2$\,s) \\
    Contact gating      & grad. threshold ($\delta{=}0.5$\,N) & None                   & None \\
    Added params        & ${\approx}2.2$\,M                & ${\approx}45$\,M       & ${\approx}2.1$\,M \\
    \bottomrule
  \end{tabular}}
\end{table}

Table~\ref{tab:arch_comparison} summarises the architectural differences between
\method and the two re-implemented baselines. All three are built on the same
$\pi_{0.5}$ backbone with identical LoRA adapters (${\approx}50$\,M trainable
parameters), output delta end-effector pose actions, and are executed by the controller of Appendix~\ref{app:controller}. ``Added
parameters'' counts only the modules introduced on top of this shared base.

\paragraph{Where the added parameters live.}
The three methods sit at very different points on the capacity-vs.-temporal-context spectrum.
\tavla\ adds a 2-layer force MLP ($6{\to}d{\to}d$, ${\approx}2.1$\,M) that
encodes a history of force readings into a single token appended to the
action-expert prefix.
\forcevla\ projects the current F/T reading through a single linear layer
($6{\to}d$, ${\approx}14$\,K) and routes it through a 4-expert LIMoE block
placed between the VLM prefix and the action expert, with almost all of its
${\approx}45$\,M added parameters sitting in the LIMoE experts and gate rather than
in the force encoder itself.
\method splits its ${\approx}2.2$\,M budget between a shared causal TCN
(${\approx}0.2$\,M, 4 dilated blocks, hidden width 64) that ingests both the
current 27-sample window and the $H{=}30$ past windows, and an AdaRMS
conditioning head (${\approx}2.1$\,M, two MLP layers and a zero-initialized
$\gamma$-projection) that modulates the action-expert RMSNorm scales at every
layer.

\paragraph{Force signal.}
All three methods consume the same 6-D wrist F/T readings from the Bota SensONE, but differ in temporal context.
\forcevla\ uses only the single reading synchronized with the current vision and proprioception step.
\tavla\ keeps a history of such readings, encoding ${\approx}2$\,s of past F/T into a single token.
\method augments the current 27-sample window with $H{=}30$ past window summaries (${\approx}2$\,s of history at 15\,Hz), which the shared causal TCN processes jointly.

\paragraph{Action space.}
For a controlled comparison, all three methods are trained with delta end-effector pose actions executed by the controller of Appendix~\ref{app:controller}, keeping the action interface and controller identical across all methods.

\section{Per-Task End-Effector Deltas and Force Profiles}
\label{app:ee_profiles}

\begin{figure}[p]
  \centering
  \captionsetup[subfigure]{skip=1pt}
  \begin{subfigure}{\linewidth}\centering
    \input{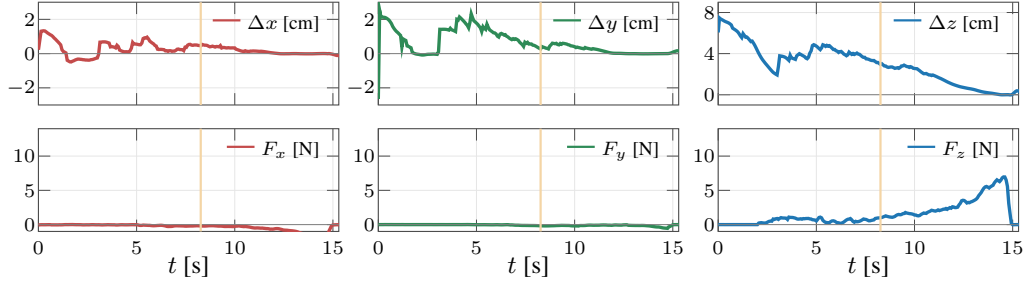}
    \subcaption{Plug insertion}\label{fig:ee_plug}
  \end{subfigure}\vspace{1mm}
  \begin{subfigure}{\linewidth}\centering
    \input{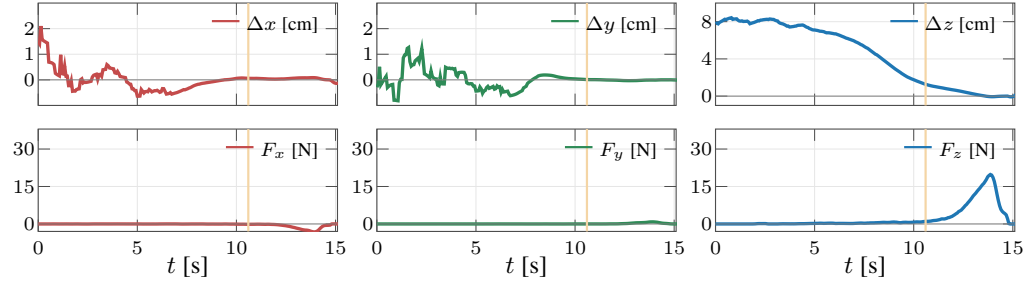}
    \subcaption{USB insertion}\label{fig:ee_usb}
  \end{subfigure}\vspace{1mm}
  \begin{subfigure}{\linewidth}\centering
    \input{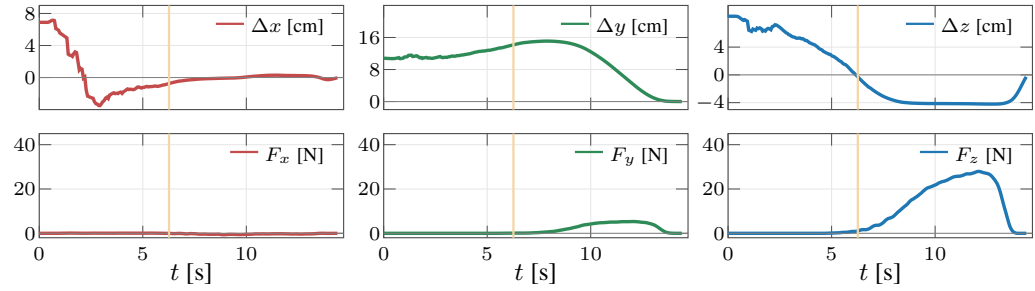}
    \subcaption{Board erasing}\label{fig:ee_board}
  \end{subfigure}\vspace{1mm}
  \begin{subfigure}{\linewidth}\centering
    \input{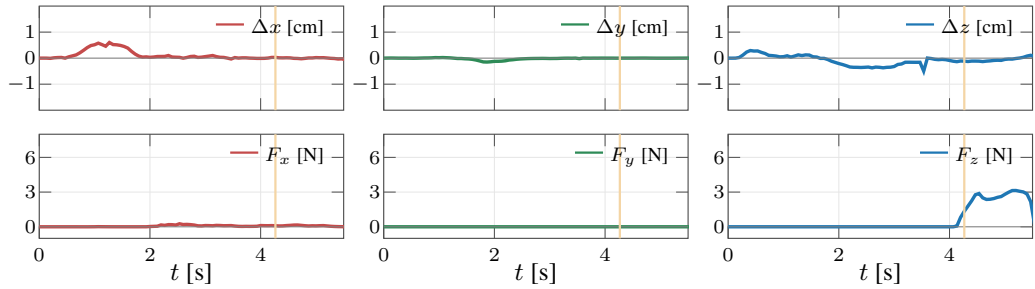}
    \subcaption{Button push}\label{fig:ee_button}
  \end{subfigure}\vspace{1mm}
  \begin{subfigure}{\linewidth}\centering
    \input{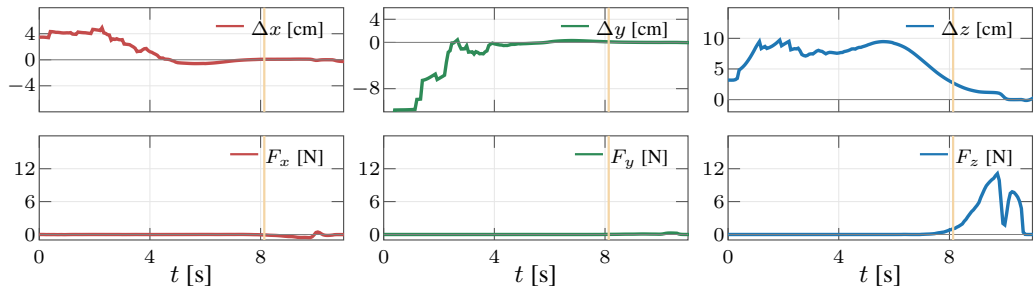}
    \subcaption{Key insertion}\label{fig:ee_key}
  \end{subfigure}
  \caption{Sample end-effector position deltas (top row) and contact
    forces (bottom row) for each task. The vertical line marks the onset of the contact phase.}
  \label{fig:ee_profiles}
\end{figure}

End-effector position deltas and contact forces for each task are included in Figure~\ref{fig:ee_profiles}. These profiles illustrate the temporal structure of the contact phase for each task, which \method's architecture is designed to capture. In particular, the transition into contact is marked by shrinking position deltas as the end-effector approaches the target and rising contact forces as it makes contact. The vertical line in each plot marks the onset of the contact phase.

\clearpage
\section{Action Prediction Accuracy}
\label{app:diagnostics}

We probe the learned flow field on 20 plug demonstrations reserved for evaluation. Given demonstrated actions $\mathbf{a}^0$, construct $\mathbf{a}^{\tau}=(1-\tau)\mathbf{a}^0+\tau\boldsymbol{\epsilon}$ with $\boldsymbol{\epsilon}\sim\mathcal{N}(0,I)$ at $\tau\in\{1.0,0.9,\ldots,0.1\}$. We compare policies adapted with LN and Beta schedules using supervised flow error $\|v_\theta(\mathbf{o},\mathbf{a}^{\tau},\tau)-(\boldsymbol{\epsilon}-\mathbf{a}^0)\|_F^2$. For $<5$\,mm actions, LN yields 8 to 13\% lower error throughout the region of low noise levels. Starting from the same noisy actions, integrate the remaining Euler trajectory and measure Cartesian reconstruction error $\|\hat{\mathbf{a}}^0-\mathbf{a}^0\|_2$. LN reduces reconstruction error at every tested start time by 5 to 13\% for $<5$\,mm actions and by 2 to 7\% for 12 to 20\,mm actions. These diagnostics support improved prediction of small corrective actions.

\section{Statistical Tests}
\label{app:statistics}
For Tables~\ref{tab:results} and~\ref{tab:force_noise}, we form a $2\times2$ contingency table of successful and failed rollouts, using 40 rollouts per method on each task. We use directional Fisher exact tests. In Table~\ref{tab:results}, the alternative is that the evaluated method has a higher success probability than $\pi_{0.5}$. In Table~\ref{tab:force_noise}, the alternative is that measured F/T yields higher success than noise substitution within each method. We report unadjusted $p$ values for individual comparisons and use a significance threshold of $\alpha=.05$.

\end{document}